\pdfoutput=1

\documentclass[11pt, dvipsnames]{article}

\usepackage{tabularx}
\usepackage[table]{xcolor}
\usepackage[final]{acl}

\usepackage{graphicx}
\usepackage{amsmath}

\usepackage{subcaption}

\usepackage[most]{tcolorbox}

\tcbset{
  figbox/.style={
    colback=white,
    colframe=grey!60,
    boxrule=0.5pt,
    arc=1pt,
    left=3pt,
    right=3pt,
    top=3pt,
    bottom=3pt
  }
}

\tcbset{
  imgbox/.style={
    colback=white,          
    colframe=gray!60,       
    boxrule=0.4pt,          
    arc=0pt,                
    left=1pt,
    right=1pt,
    top=1pt,
    bottom=1pt
  }
}

\usepackage{enumitem} 
\tcbuselibrary{breakable}
\usepackage{booktabs}
\usepackage{siunitx}
\usepackage{adjustbox}
\usepackage{subcaption}
\usepackage[T1]{fontenc}
\usepackage[utf8]{inputenc}
\usepackage{amsmath}
\usepackage{booktabs}
\usepackage{graphicx}
\usepackage{xcolor}
\usepackage{tikz}
\usepackage{booktabs}
\usepackage{adjustbox}
\usepackage{array,multirow,graphicx}
\usepackage{float}
\usepackage{makecell}
\usepackage{amssymb}
\usepackage{pifont}

\newcommand{\cnerabbr}[1]{\textsc{#1}} 

\newcommand{\lightrule}{\specialrule{0.3pt}{0pt}{0pt}}

\NewDocumentCommand{\corefRed}{m m}{{\textcolor{Red}{\textbf{[}}#1\textcolor{Red}{\textbf{]}\textsubscript{#2}}}}
\NewDocumentCommand{\corefGreen}{m m}{{\textcolor{Green}{\textbf{[}}#1\textcolor{Green}{\textbf{]}\textsubscript{#2}}}}

\NewDocumentCommand{\corefBlue}{m m}{{\textcolor{Blue}{\textbf{[}}#1\textcolor{Blue}{\textbf{]}\textsubscript{#2}}}}
\NewDocumentCommand{\corefPurple}{m m}{{\textcolor{Purple}{\textbf{[}}#1\textcolor{Purple}{\textbf{]}\textsubscript{#2}}}}
\NewDocumentCommand{\corefBlack}{m m}{{\textcolor{Black}{\textbf{[}}#1\textcolor{Black}{\textbf{]}\textsubscript{#2}}}}
\NewDocumentCommand{\corefOlive}{m m}{{\textcolor{olive}{\textbf{[}}#1\textcolor{olive}{\textbf{]}\textsubscript{#2}}}}

\newcommand{\CNERART}{\cnerabbr{ARTF}}     
\newcommand{\CNERRELA}{\cnerabbr{REL}}     
\newcommand{\CNERCULT}{\cnerabbr{CULT}}    
\newcommand{\CNERDISE}{\cnerabbr{DIS}}     

\usepackage{algorithm}
\usepackage{algorithmic}
\usepackage{hyperref}
\usepackage{multirow}
\usepackage{multicol}
\usepackage{graphics}
\usepackage{array}
\usepackage{array, diagbox}

\usepackage{booktabs}
\usepackage{arydshln}
\usepackage{multirow}
\usepackage{times}
\usepackage{latexsym}
\usepackage[multiple]{footmisc}
\definecolor{rowgray}{gray}{0.95}

\usepackage{times}
\usepackage{latexsym}

\usepackage[T1]{fontenc}

\usepackage[utf8]{inputenc}

\usepackage{microtype}

\usepackage{inconsolata}

\usepackage{xcolor}

\usepackage{tikz}
\usetikzlibrary{arrows.meta,positioning,fit,calc}

\title{Interpretable Coreference Resolution Evaluation Using Explicit Semantics}

\author{Bruno Gatti$^{1,\dagger}$, Giuliano Martinelli$^{1,\dagger}$, Roberto Navigli$^{1,2,\dagger}$\\
$^1$Sapienza NLP Group, Sapienza University of Rome\\
$^2$Babelscape\\
\texttt{\{gatti, martinelli, navigli\}@diag.uniroma1.it}}

\begin{document}
\maketitle

\begingroup
\renewcommand\thefootnote{}\footnotetext{$\dagger$ Authors contributed equally.}
\endgroup


\begin{abstract}
Coreference resolution is typically evaluated using aggregate statistical metrics such as CoNLL-F$_1$, which measure structural overlap between predicted and gold clusters. 
While widely used, these metrics offer limited diagnostic insights, penalizing errors without revealing whether a system struggles with specific semantic categories, such as people, locations, or events, and making it difficult to interpret model capabilities or derive actionable improvements.
We address this gap by introducing a semantically-enhanced evaluation framework for coreference resolution. 
Our approach overlays Concept and Named Entity Recognition (CNER) onto coreference outputs, assigning semantic labels to nominal mentions and propagating them to entire coreference clusters.
This enables the computation of typed scores aimed at evaluating mention extraction and linking capabilities stratified by semantic class.
Across our experiments on OntoNotes, LitBank, and PreCo, we show that our framework uncovers systematic weaknesses that remain obscured by aggregate metrics. 
Furthermore, we demonstrate that these diagnostics can be used to design targeted, low-cost data augmentation strategies, achieving measurable out-of-domain improvements.
\end{abstract}

\section{Introduction}
While coreference resolution models have improved substantially in recent years, evaluation methodologies have evolved far more slowly and continue to suffer from longstanding limitations.

\begin{figure}[t]
    \centering

    \begin{subfigure}[b]{\linewidth}
        \centering
        \begin{tcolorbox}[imgbox]
            \includegraphics[width=\linewidth]{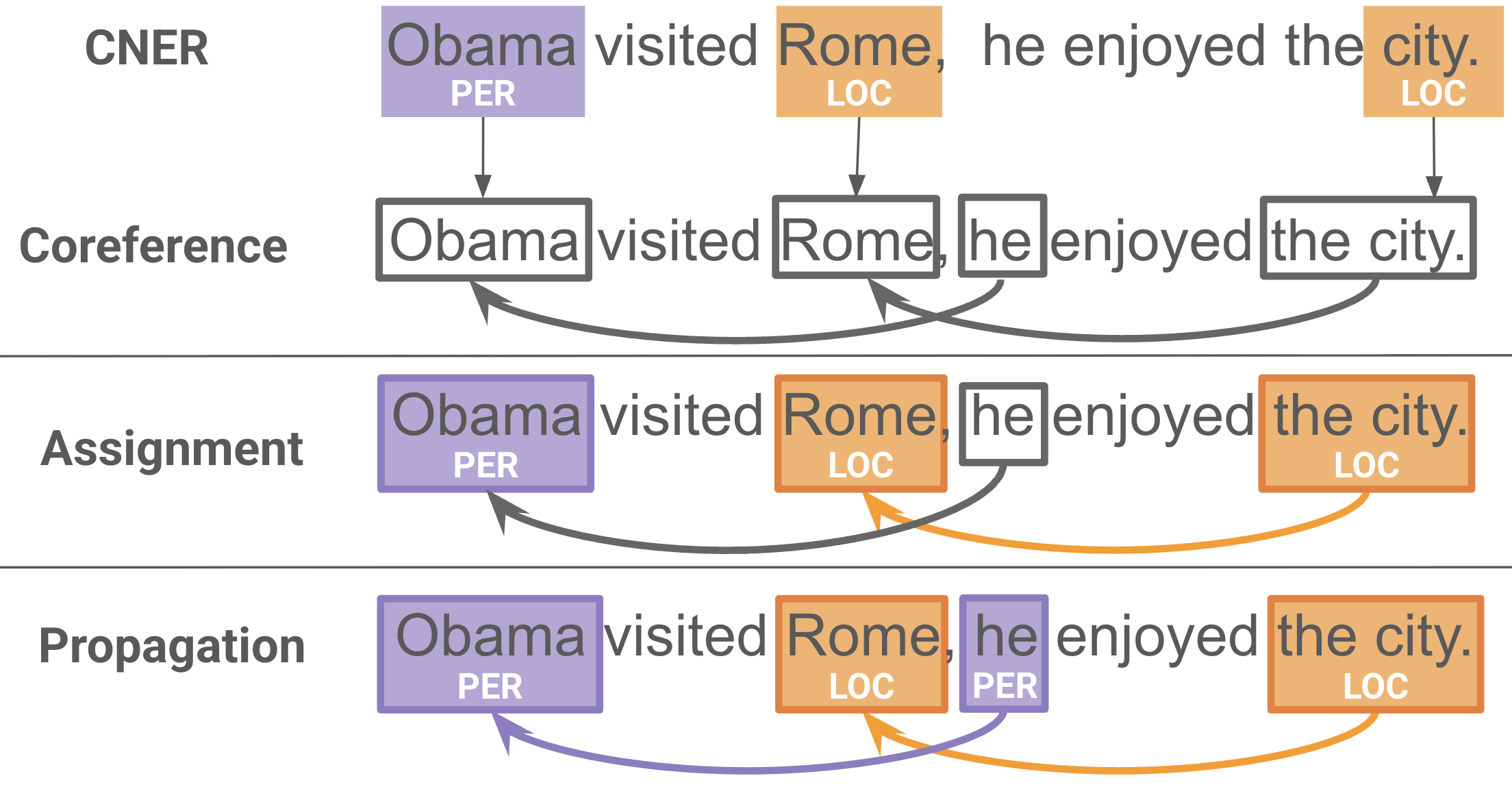}
        \end{tcolorbox}
        \caption{Our labeling and propagation technique.}
        \label{fig:three_images_a}
    \end{subfigure}

    \vspace{0.4em}

    \begin{subfigure}[b]{0.49\linewidth}
        \centering
        \begin{tcolorbox}[imgbox]
            \includegraphics[width=\linewidth]{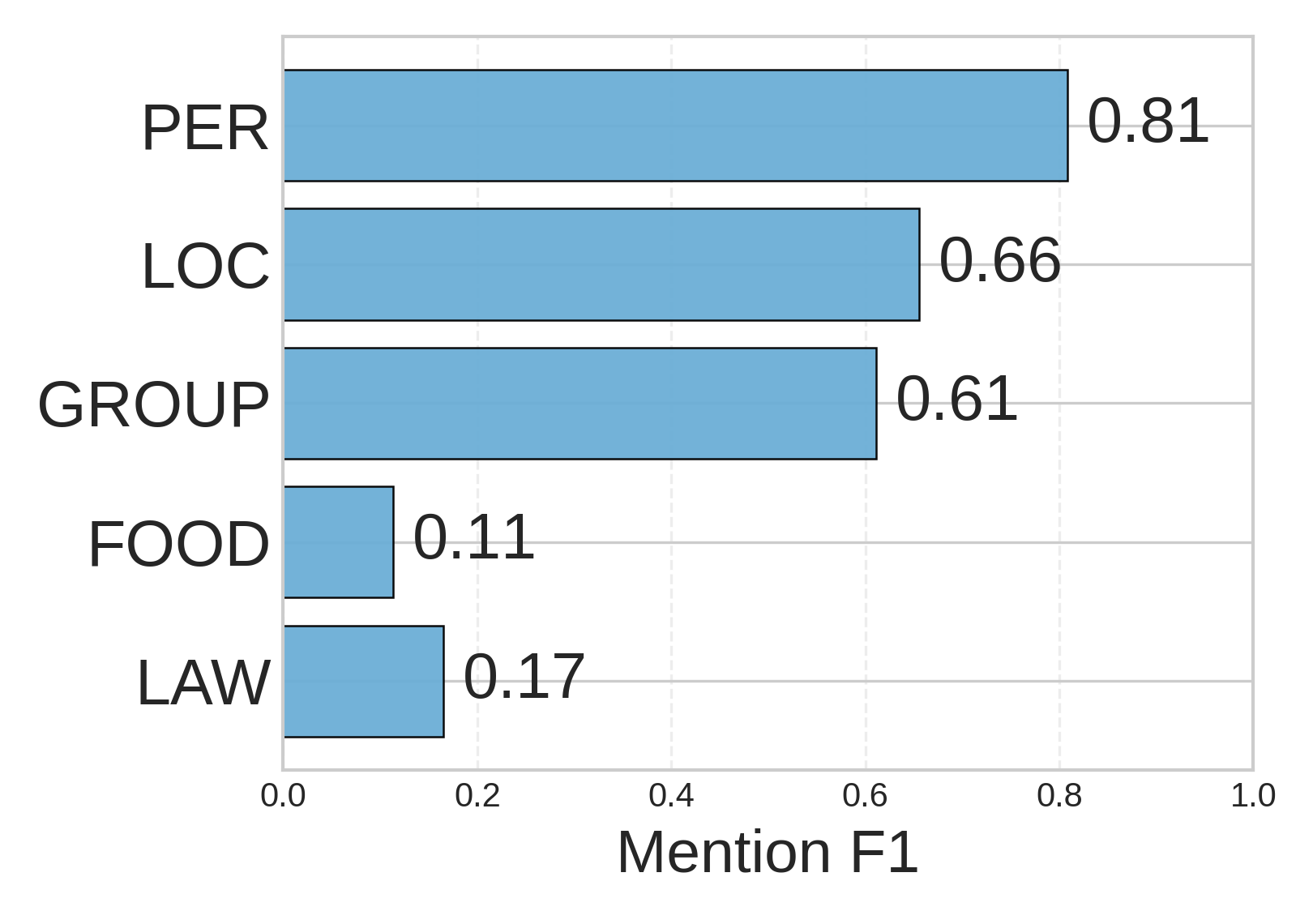}
        \end{tcolorbox}
        \caption{Semantically-enhanced evaluation.}
        \label{fig:three_images_b}
    \end{subfigure}\hfill
    \begin{subfigure}[b]{0.49\linewidth}
        \centering
        \begin{tcolorbox}[imgbox]
            \includegraphics[width=\linewidth]{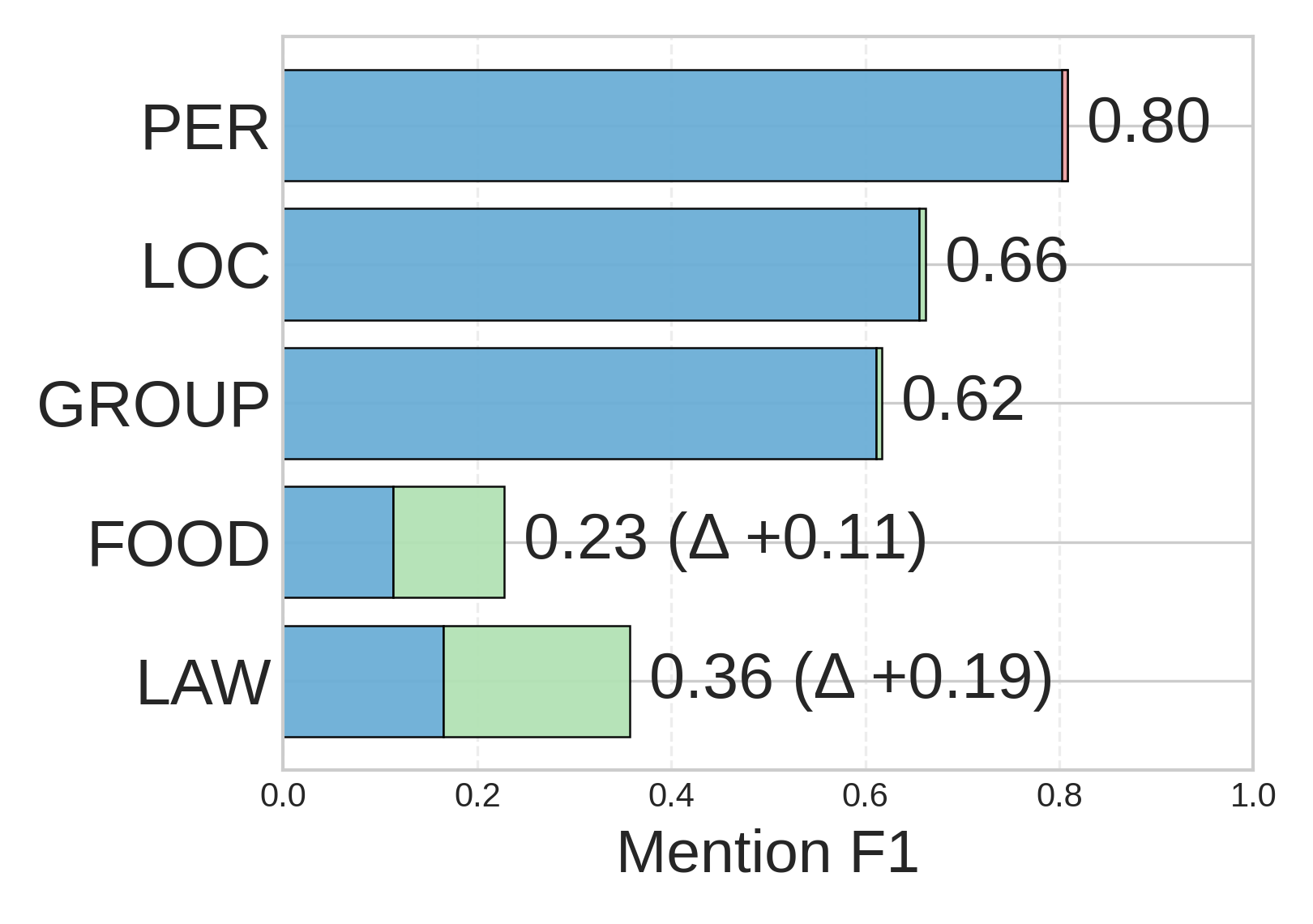}
        \end{tcolorbox}
        \caption{Targeted data augmentation.}
        \label{fig:three_images_c}
    \end{subfigure}

    \caption{Our three main contributions: (a) Our labeling and propagation technique that overlays CNER onto coreference outputs; (b) Our interpretable per-class semantic analysis of coreference capabilities; (c) The improvements of targeted data augmentation on less represented classes. }
    \label{fig:three_images}
\end{figure}

Standard metrics such as MUC \citep{vilain-etal-1995-model}, B\textsuperscript{3} \citep{bagga-baldwin-1998-entity},  
and CEAF\textsubscript{$\phi$4} \citep{luo-2005-coreference}, popularized by the CoNLL-2012 shared task \citep{pradhan-etal-2012-conll}, are based on statistical methods that focus on exact matches of mentions and coreference links.
While effective as aggregate measures, they ignore semantic and contextual information.
Therefore, minor span discrepancies or annotation mismatches can be penalized as full errors, complicating meaningful comparison across annotation schemes, genres, and domains.

A further limitation of these metrics is their lack of interpretability, as single aggregate scores often obscure systematic failure modes and conflate qualitatively different errors. 
For example, a model may perform well on person-centric narrative chains while failing on event- or object-related coreference, yet this degradation is not reflected in standard evaluation scores. 
This issue becomes especially pronounced under domain shift, where models may degrade unevenly across semantic categories without any clear diagnostic signal.
Prior work has attempted to improve the interpretability of coreference evaluation by incorporating explicit semantic information \cite{agarwal-etal-2019-evaluation}.
This approach is, however, constrained by the limited scope of the conventional Named Entity Recognition (NER) tagset, which only covers named entities and excludes common nouns, despite the fact that nominal concepts constitute a large fraction of coreference mentions.
For this reason, NER-based evaluation can only sparsely annotate coreference clusters and is restricted to a small set of coarse-grained categories (typically \textsc{PER}, \textsc{ORG}, \textsc{LOC}, and \textsc{MISC}), limiting meaningful analysis across richer semantic distinctions.

To address these limitations, we propose a new evaluation framework that adopts a novel two-step labeling and propagation technique to tag coreference clusters using Concept and Named Entity Recognition \citep[CNER]{martinelli-etal-2024-cner} as a semantic layer. 
Unlike conventional NER systems, CNER assigns semantic categories to both named entities  (e.g., \emph{Obama}, \emph{Rome}, \emph{Moby Dick}) and nominal concepts  (e.g., \emph{president}, \emph{city}, \emph{whale}) within a unified label inventory, ensuring that a meaningful semantic label can be assigned to the large majority of coreference mentions.
As illustrated in Figure~\ref{fig:three_images}, we overlay CNER annotations onto coreference outputs using a simple overlap-based alignment, followed by cluster-level label propagation.
This enables the computation of typed Mention and Link F$_1$ scores, turning coreference evaluation into a semantically-aware diagnostic interface. 
Beyond improved interpretability, we also show that these measures can be leveraged to design targeted and low-cost data augmentation strategies yielding measurable improvements in out-of-domain generalization. Specifically, our contributions are as follows:

\begin{enumerate}
    \item \textbf{A labeling and propagation technique for semantically-enhanced coreference evaluation}, based on a simple two-step approach that merges CNER tags with coreference outputs, providing dense and granular semantic annotations.
    \item \textbf{A fine-grained analysis of coreference models and datasets}, revealing systematic, semantically interpretable error patterns that remain hidden under standard metrics.
    \item \textbf{An actionable evaluation framework}, demonstrating how our novel diagnostics can guide targeted and cost-effective data augmentation, leading to improved out-of-domain performance.
\end{enumerate}

\noindent
We release the code of our evaluation framework and the data used at \url{https://github.com/SapienzaNLP/cner-coref}.

\section{Related Work}

\subsection{Coreference Resolution Evaluation}
In recent years, coreference resolution has typically been evaluated using MUC \citep{vilain-etal-1995-model}, B\textsuperscript{3} \citep{bagga-baldwin-1998-entity},  
and CEAF\textsubscript{$\phi$4} metrics \citep{luo-2005-coreference}, later combined into the CoNLL score \citep{pradhan-etal-2012-conll}.  

While these metrics provide stable aggregate performance estimates, they offer little diagnostic insight: they cannot reveal whether a model systematically fails on particular semantic categories or mention types, and they often obscure qualitatively distinct error patterns. 

To address structural biases, \citet{moosavi-strube-2016-coreference} propose the LEA metric to better weight entity importance, while \citet{moosavi-etal-2019-using} introduce MINA algorithms to decouple the evaluation of coreference logic from strict boundary detection. 
Parallel efforts focus on diagnosing specific failures through algorithmic categorization; \citet{kummerfeld-klein-2013-error} develop a toolkit to automatically group errors into intuitive types based on entity transformations, revealing that many persistent failures stem from semantic or discourse mismatches invisible to link-based metrics. 
From a different methodological perspective, \citet{martschat-etal-2015-analyzing} introduce a framework that uses spanning trees to represent coreference structures, allowing for a systematic analysis of precision and recall errors.

Despite the availability of these analytical tools, recent literature suggests that evaluation challenges are deeply tied to the underlying datasets' annotation guidelines. 
\citet{porada-etal-2024-challenges} demonstrate that perceived failures in out-of-domain generalization are frequently artifacts of inconsistent mention definitions rather than genuine linguistic shortcomings. 
This echoes arguments by \citet{DBLP:journals/corr/abs-2112-09742}, who contends that the field's heavy reliance on the OntoNotes schema has artificially constrained the task's definition. 
A prominent example of this constraint is the assumption of strict coreference; while psycholinguistic evidence demonstrates that coreference often involves a more nuanced "near-identity" rather than a strict binary classification \citep{recasens-etal-2013-linguistic}, standard datasets enforce strict-identity annotations. 


However, coreference evaluation continues to rely on aggregate statistical scores that lack fine-grained, semantically informed diagnostics, which is the gap our work aims to address.

\subsection{Semantics and Coreference Models}
A substantial body of prior work has explored incorporating semantic information within coreference models.
Early feature-based systems leverage semantic class information and lexical
relations to constrain antecedent selection \citep{ng-2007-semantic}.
Subsequent joint models integrate typing, linking, and coreference into unified architectures \citep{durrett-klein-2014-joint, chen-etal-2017-robust, agarwal-etal-2022-entity},
demonstrating that shared semantic representations can improve cluster coherence.
Neural approaches further refine this paradigm by injecting type information
directly into span- or entity-level representations
\citep{clark-manning-2015-entity, khosla-rose-2020-using}.
More recently, \citet{mtumbuka-schockaert-2024-encore} show that coreference structure itself can act as supervision for fine-grained entity typing.
Differently, \citet{agarwal-etal-2019-evaluation} focus on using semantics as an evaluation lens for coreference behavior, analyzing coreference performance by semantic type using a standard NER-style tagset.
However, this approach comes with two limits: first, only a small set of coarse-grained
categories is considered (typically \textsc{PER}, \textsc{ORG}, \textsc{LOC} and \textsc{MISC}); second, the majority of nominal, abstract, and conceptual mentions are left untyped.
The resulting evaluation offers a shallow diagnostic view that is unable to reveal category-specific failure modes beyond named entities.
Our work addresses this limitation by enabling semantic evaluation at a much finer granularity, covering both entities and concepts, without altering the coreference model itself.
To this end, we employ Concept and Named Entity Recognition \citep[CNER]{martinelli-etal-2024-cner}, which is introduced specifically to address the narrow scope of traditional NER.  
CNER brings together named entities and nominal concepts under a unified tagset with 29 categories, ranging from \textsc{PERSON} and \textsc{LOCATION} to \textsc{RELATION}, \textsc{EVENT}, \textsc{PLANT}, and \textsc{SUPERNATURAL}, among others.
We detail the full inventory in Appendix ~\ref{app:cner}. 
This yields not only a much denser annotation layer -- since nearly every mention in a coreference chain can be mapped to a precise CNER category -- but also far broader semantic granularity than standard NER.

\section{Labeling and Propagation Technique} 
In this Section, we introduce our two-step labeling and propagation technique that overlays coreference outputs with the semantic annotations produced by CNER.
This simple-yet-effective method assigns a semantic category to each coreference cluster by first aligning mentions to labeled spans using an overlap-based criterion, and then propagating these labels at the cluster level via majority voting.

\subsection{Formal Overview}
Let $D$ be a document for which a coreference model has predicted a set of coreference mentions
$M = \{ m_1, m_2, \dots, m_n \}$ and clusters $\mathcal{G}$, where each cluster $G \in \mathcal{G}$ contains a set of  coreferential mentions $m^{G}_1, m^{G}_2, \dots, m^{G}_l$.
For $D$ we compute a set of CNER-annotated spans $C = \{ c_1, c_2, \dots, c_k \}$ with semantic labels $\mathcal{L}(c_j) \in \mathcal{T}$, where $\mathcal{T}$ is the inventory of CNER categories presented in Appendix~\ref{app:cner}, Table \ref{tab:cner-categories}.
The goal of our technique is to assign to each coreference cluster $G$ a semantic label $\mathcal{S}(G) \in \mathcal{T}$.
To do so, we base our approach on two steps: i) Mention Assignment, in which we adopt a span overlap-based technique to assign a CNER label to all nominal mentions, and ii) Category Propagation, in which we assign a label to each coreference cluster through majority voting and propagate the labels to all the mentions in the cluster, including pronouns.

\subsection{Mention Assignment}
\label{mention_assignment}
In the first step, we attempt to directly assign a semantic label to each mention.
To quantify the alignment between a mention $m_i$ and a CNER span $c_j$,
we define an overlap function $\Omega$ based on token-level Jaccard similarity:
\[
\Omega(m_i, c_j) =
\frac{| \text{span}(m_i) \cap \text{span}(c_j) |}
     {| \text{span}(m_i) \cup \text{span}(c_j) |}
\]
where $\text{span}(\cdot)$ denotes the set of indices of the words encompassed by mention $m_i$ or CNER tag $c_j$.
For a given mention $m_i$, we select the CNER span $\hat{c}_j$ with the highest overlap score.
If this maximum score exceeds a threshold $\tau$ (set to $0.5$ in all experiments), the mention $m_i$ is assigned the corresponding label $l_i = \mathcal{L}(\hat{c}_j)$.
Mentions that do not sufficiently overlap with any CNER span are left unlabeled.

\subsection{Category Propagation}
In the second step, we assign a semantic label to each coreference cluster based on the labels of its mentions, and then propagate this label to all unlabeled mentions in the cluster.
For each cluster $G \in \mathcal{G}$ that contains at least one labeled mention, we compute its cluster-level label via majority voting.
Specifically, for each semantic category $t \in \mathcal{T}$, we count the number of mentions in $G$ labeled as $t$, and select the most frequent category:
\[
\mathcal{S}(G) =
\arg\max_{t \in \mathcal{T}}
\left| \{ m^{G} \in G \mid \mathcal{L}(m^{G}) = t \} \right|.
\]
If two or more labels occur with equal frequency, we break ties by selecting the label whose mentions have, on average,  the highest overlap $\Omega$ with their corresponding CNER spans.  
Once the dominant label $\mathcal{S}(G)$ is determined, it is propagated to all mentions in $m^{G} \in G$, also including pronominal and other unlabeled mentions.
In the particular edge case where coreference clusters contain no nominal mentions, our technique cannot assign tags; however, as shown in Section \ref{results}, this occurs infrequently due to the high density of CNER annotations.

\section{Semantic Evaluation Framework}
\label{sec:typed-eval}

We now present our semantic framework for interpretable and actionable evaluation of coreference resolution. 
After propagation, each cluster (thus, each mention) is associated with a CNER type.
We use these types to report coreference performance stratified by semantic class via two typed F$_1$ scores, i.e., Mention F$_1$ and Link F$_1$.

\paragraph{Mention F$_1$ Score}
The Mention F$_1$ score quantifies, for a given class $t \in \mathcal{T}$, the quality of mention extraction independently of clustering.  
A predicted mention $m$ with label $\mathcal{L}(m) = t$ is treated as a true positive if $m$ is a gold-standard mention.
It is defined as a traditional F$_1$ score, i.e., the harmonic mean of mention precision and recall, where precision measures the proportion of predicted mentions that exactly match gold-standard mentions, and recall measures the proportion of gold mentions that are correctly predicted.
\paragraph{Link F$_1$ Score}
\label{par:linking_score}
The Link F$_1$ score evaluates the quality of predicted coreference links between mentions, i.e., pairs of coreference mentions that belong to the same cluster.  
A predicted link $p^G = \{m_1^G, m_2^G \}$ between two mentions in the same cluster $G$ with label $\mathcal{S}(G) = t$ is treated as a true positive for class $t$ if $p^G$ is present among the gold links.  
Unlike Mention F$_1$, which isolates span detection, Link F$_1$ captures the structural quality of clustering.  
In our experiments, we compute Link F$_1$ using gold mentions to control for mention detection effects and to focus exclusively on the model’s linking performance.

\paragraph{Metrics Interpretation and Downstream Usage}
Together, Mention F$_1$ and Link F$_1$ provide a coherent and interpretable decomposition of coreference performance. 
Mention F$_1$ captures \textit{what} the model recognizes as entity mentions among those annotated in the dataset, while Link F$_1$ captures \emph{how} these mentions are connected into referential structures.  
When computed per semantic category, these metrics provide fine-grained diagnostics; we use them in Section~\ref{sec:rq3} to guide targeted data augmentation toward underperforming categories.

\section{Experimental Setup}
\label{sec:exp-setup}



\subsection{Datasets}

We conduct experiments on three English coreference resolution datasets that differ substantially in domain, annotation scope, and semantic coverage.

\noindent
\textbf{OntoNotes} \citep{pradhan-etal-2012-conll} is a large, multi-genre corpus released as part of the CoNLL-2012 shared task on multilingual coreference resolution. Its diverse genre and variety of noun types (proper nouns, common-noun mentions, pronominal mentions) cover a broad and balanced distribution of entity types.

\noindent
\textbf{LitBank} \citep{bamman-etal-2020-annotated} is a coreference corpus derived from literary texts, annotated under restrictive guidelines that include only six entity types: \textsc{PER} (persons), \textsc{FAC} (facilities), \textsc{LOC} (locations), \textsc{GPE} (geopolitical entities), \textsc{ORG} (organizations), and \textsc{VEH} (vehicles).
According to the original report, \textsc{PER} mentions account for 83.1\% of all mentions, followed by \textsc{FAC} (8.0\%), \textsc{LOC} (4.4\%), \textsc{GPE} (3.3\%), \textsc{VEH} (0.7\%), and \textsc{ORG} (0.5\%).

\noindent
\textbf{PreCo} \citep{chen-etal-2018-preco} is a large-scale English coreference dataset comprising terminology mostly coming from preschoolers’ vocabulary, which emphasizes entities in everyday contexts.
\begin{figure}[t]
    \centering
    \includegraphics[width=1\linewidth]{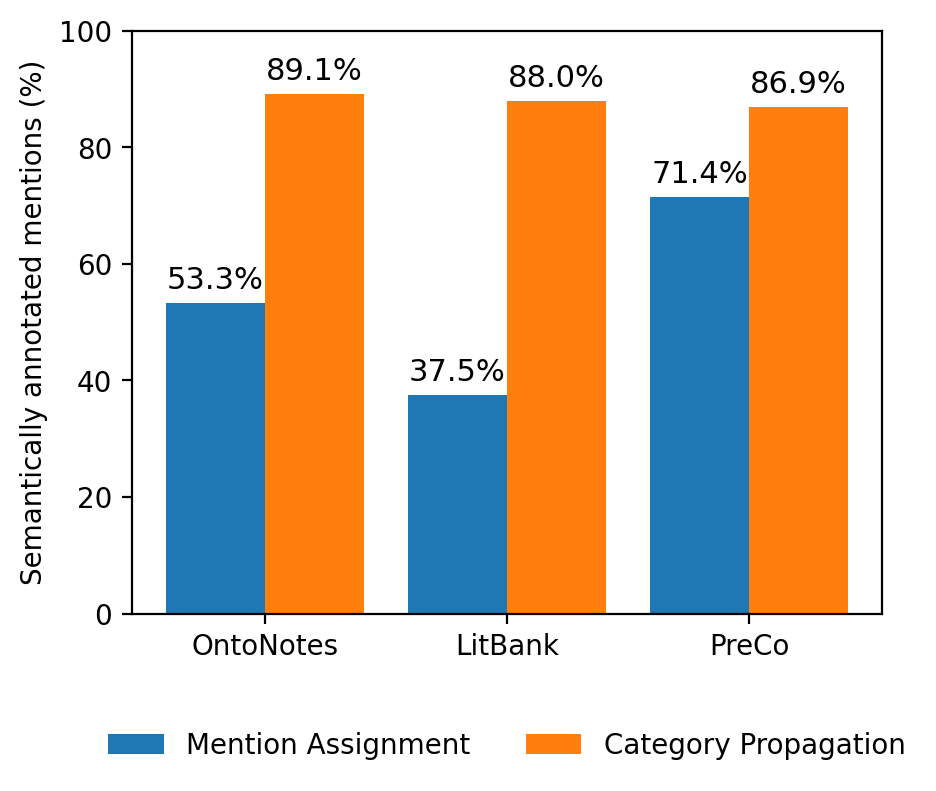}
    \caption{Percentage of CNER-annotated coreference mentions after Mention Assignment (in blue) and after Category Propagation (in orange).}
    \label{fig:coverage-propagation}
\end{figure}


\begin{table}
\centering
\small
\setlength{\tabcolsep}{6pt}
\renewcommand{\arraystretch}{1.05}

\begin{adjustbox}{max width=\columnwidth}
\begin{tabular}{
l r
@{\hspace{2.2em}}
c
@{\hspace{2.2em}}
l r
}
\toprule
\multicolumn{2}{c}{\textbf{WikiNEuRal (NER)}} &
&
\multicolumn{2}{c}{\textbf{CNER}} \\
\cmidrule(lr){1-2}\cmidrule(lr){4-5}
Type & Support & & Type & Support \\
\midrule
\textsc{PER}  & 46{,}147 & & \textsc{PER}   & 58{,}153 \\
\textsc{LOC}  & 12{,}988 & & \textsc{LOC}   & 12{,}458 \\
\textsc{ORG}  & 9{,}756  & & \textsc{ORG}   & 8{,}645 \\
\addlinespace

\multirow{10}{*}{\textsc{MISC}} &
\multirow{10}{*}{7{,}096} &
\multirow{10}{*}{\(\left\{\vphantom{\begin{matrix}
\textsc{GROUP}\\
\textsc{MEDIA}\\
\textsc{SUPER}\\
\CNERART\\
\textsc{EVENT}\\
\vdots\\
\CNERRELA\\
\CNERCULT\\
\CNERDISE
\end{matrix}}\right.\)} &
\textsc{GROUP} & 5{,}222 \\
& & & \textsc{MEDIA} & 4{,}034 \\
& & & \textsc{SUPER} & 3{,}782 \\
& & & \CNERART      & 2{,}291 \\
& & & \textsc{EVENT} & 1{,}091 \\
& & & \multicolumn{2}{c}{\emph{(\dots)}} \\
& & & \CNERRELA & 15 \\
& & & \CNERCULT & 15 \\
& & & \CNERDISE & 10 \\
& & & \textsc{PLANT} & 5 \\

\bottomrule

\end{tabular}
\end{adjustbox}

\caption{Distribution comparison of semantic types and supports between NER and CNER on OntoNotes.}
\label{tab:ner-vs-cner-adjacent}
\end{table}


\begin{figure}[t!]
    \centering
    \includegraphics[width=1\linewidth]{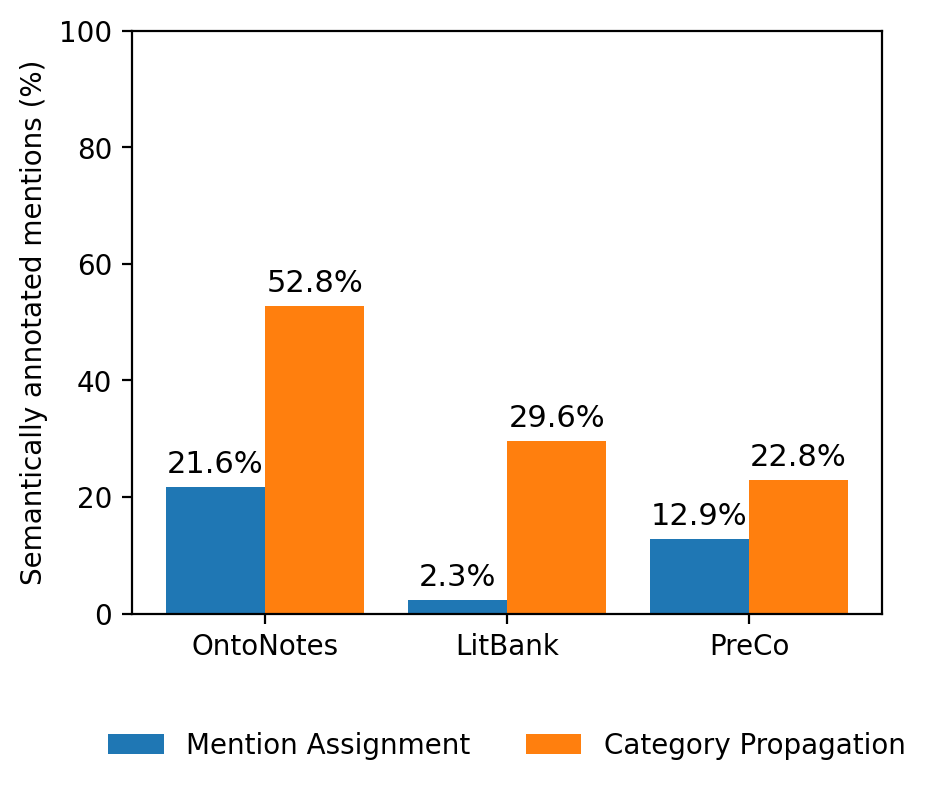}
    \caption{Percentage of NER-annotated coreference mentions after Mention Assignment (in blue) and after Category Propagation (in orange).}
    \label{tab:cner-vs-ner-coverage}
\end{figure}


\subsection{Comparison Systems}
\label{comparison_systems}

\subsubsection{Coreference Models}
To isolate the effects of domain and semantic distribution, we evaluate models with the same underlying coreference architecture trained on different datasets.
This checks for architectural confounds and allows us to focus on out-of-domain behavior.
As the underlying coreference model, we use Maverick \cite{martinelli-etal-2024-maverick} in its multi-expert scorers version, a recent encoder-only neural architecture that jointly models mention extraction and clustering, achieving state-of-the-art performance while remaining computationally efficient.
We consider three Maverick variants from the official repository\footnote{\href{https://hf.co/collections/sapienzanlp/maverick-coreference-resolution}{Maverick official github repository}}: \textbf{maverick-mes-ontonotes}, \textbf{maverick-mes-litbank}, \textbf{maverick-mes-preco}, trained respectively on OntoNotes, LitBank and PreCo.


\begin{table*}[ht]
\centering
\small
\begin{tabular}{l ccc ccc}
\toprule
 & \multicolumn{3}{c}{\textbf{Mention F$_{1}$}}
 & \multicolumn{3}{c}{\textbf{Link F$_{1}$}}\\
\cmidrule(lr){2-4} \cmidrule(lr){5-7}
\textbf{Model} 
 & \textbf{OntoNotes} & \textbf{LitBank} & \textbf{PreCo}
 & \textbf{OntoNotes} & \textbf{LitBank} & \textbf{PreCo} \\
\midrule
maverick-mes-ontonotes & \textbf{0.85} & 0.48 & 0.40 & \textbf{0.77} & \textbf{0.53} & 0.57 \\
maverick-mes-litbank   & 0.40 & \textbf{0.78} & 0.31 & 0.43 & \textbf{0.53}  & 0.47 \\
maverick-mes-preco     & 0.53 & 0.35 & \textbf{0.93} & 0.47 & 0.46 & \textbf{0.82} \\
\bottomrule
\end{tabular}
\caption{Macro Mention F$_1$ and Link F$_1$ results for Maverick models across OntoNotes, LitBank, and PreCo.}
\label{tab:typed-macro}
\end{table*}


\begin{figure}[ht]
    \centering

    \begin{subfigure}{0.5\linewidth}
        \centering
        \includegraphics[width=\linewidth]{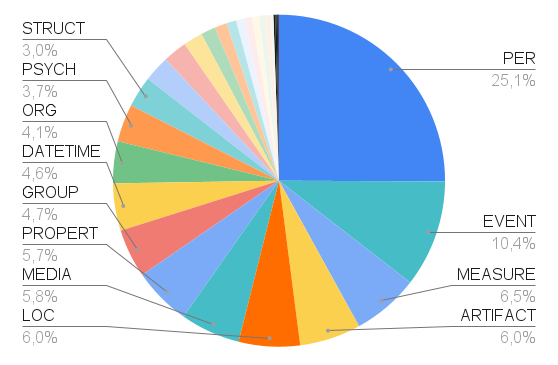}
        \caption{PreCo.}
        \label{fig:category-preco}
    \end{subfigure}\hfill
    \begin{subfigure}{0.5\linewidth}
        \centering
        \includegraphics[width=\linewidth]{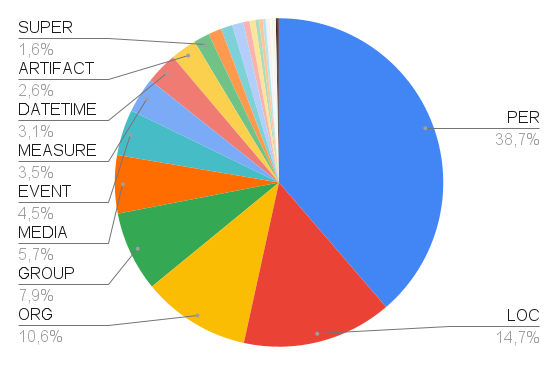}
        \caption{OntoNotes.}
        \label{fig:category-ontonotes}
    \end{subfigure}

    \vspace{0.4em}

    \begin{subfigure}{\linewidth}
        \centering
        \includegraphics[width=0.8\linewidth]{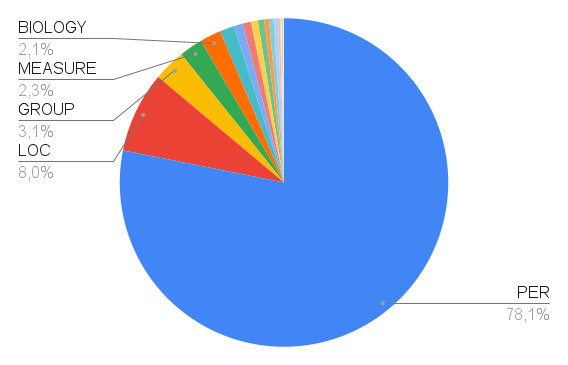}
        \caption{LitBank.}
        \label{fig:category-litbank}
    \end{subfigure}

    \caption{Distribution of propagated CNER semantic categories for gold coreference mentions.
    }
    \label{fig:category-distributions}
\end{figure}


\begin{figure*}[t]
    \centering
    \includegraphics[width=0.75\linewidth]{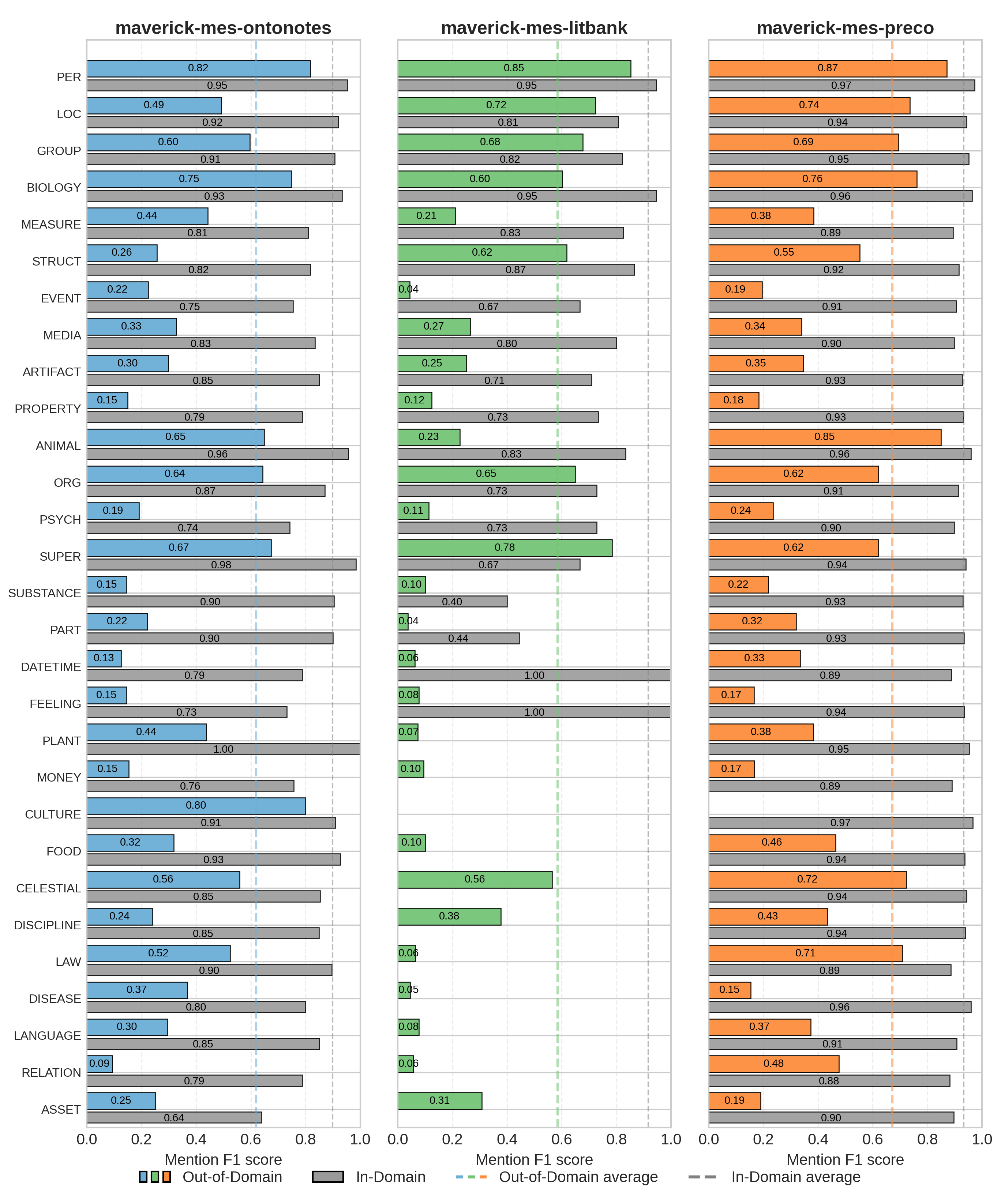}
    \caption{Per-class Mention F$_1$ scores for each model. In-domain results are shown in grey, while out-of-domain results are shown in different colors and computed as the average performance on the two datasets not used for training. Dashed vertical lines indicate the mean in-domain and out-of-domain scores. Classes are ordered by decreasing category frequency in LitBank so as to highlight out-of-domain performance.}
    \label{fig:per-class mentionF$_1$}
\end{figure*}


\subsubsection{Semantic Annotation Layer}

Our framework relies on CNER as the semantic annotation layer, and all our experiments are conducted using the official CNER checkpoints\footnote{\href{https://huggingface.co/Babelscape/cner-base}{Official CNER checkpoints}}.

To compare against prior approaches that rely on standard NER-style semantic categories, we experiment with replacing CNER with a standard NER, i.e., WikiNEuRal \citep{tedeschi-etal-2021-wikineural-combined}, a widely used NER system that predicts four traditional coarse-grained NER classes (\textsc{PER}, \textsc{LOC}, \textsc{ORG}, \textsc{MISC}).


\section{Results}
\label{results}
In Section \ref{sec:rq1}, we measure the effectiveness of our labeling and propagation technique in terms of mention coverage and include a comparison between CNER and standard NER.
Subsequently, in Section \ref{sec:rq2}, through our evaluation framework, we analyze semantic gaps in current coreference datasets and models.
Finally, in Section~\ref{sec:rq3}, we explore a possible downstream application of our evaluation to improve model results.

\subsection{Effectiveness of our technique}
\label{sec:rq1}


We evaluate the effectiveness of our labeling and propagation technique by i) measuring mention coverage with CNER, ii) comparing CNER and standard NER as semantic annotation layers, and iii) manually measuring the performance of our labeling and propagation technique on $30\%$ of the entire LitBank test set, thereby checking the correctness of CNER labels assigned to gold clusters.

\paragraph{Mention Coverage.}
Figure~\ref{fig:coverage-propagation} reports the mention coverage, i.e., the proportion of coreference mentions assigned to a CNER category after Mention Assignment and after Category Propagation.
Mention Assignment labels only nominal mentions aligned via span overlap (Section~\ref{mention_assignment}), covering between 37.5\% of mentions on LitBank and 71.4\% on PreCo. 
After Category Propagation, coverage rises to \(\sim\)90\% on all datasets.
The remaining unlabeled mentions, after inspection, are found to be mostly clusters containing only pronominal mentions, which cannot be labeled during the Mention Assignment step.
Further analysis on pronominal mentions coverage is provided in Appendix \ref{app:propagation_heuristic} in Table~\ref{tab:cner-coverage} and~\ref{tab:ner-coverage}, and in Appendix~\ref{app:propagation_examples}, where we present qualitative examples of our labeling and propagation technique.


\paragraph{Investigating NER vs CNER.}
\label{par:ner_as_a_baseline}
Since our framework is agnostic to the choice of the semantic annotation layer, we test whether standard NER can provide an alternative to CNER.
We compare CNER and NER along two dimensions: i) mention coverage, as the percentage of CNER annotated mentions, and ii) semantic granularity, i.e., the number and diversity of semantic categories available for analysis.
Regarding mention coverage, Figure~\ref{tab:cner-vs-ner-coverage} shows the percentage of annotated mentions by NER during our two-step labeling and propagation technique.
Compared to CNER, NER assigns semantic labels to substantially fewer mentions across all datasets.
Even after propagation, NER coverage remains limited (52.8\% on OntoNotes, 29.6\% on LitBank, and 22.8\% on PreCo), making CNER more suited as an underlying semantic layer.

Regarding semantic granularity, Table~\ref{tab:ner-vs-cner-adjacent} shows that NER-based evaluation collapses model behavior into only four coarse categories (\textsc{PER}, \textsc{LOC}, \textsc{ORG}, \textsc{MISC}), each with high support and limited differentiation, providing low diagnostic resolution.
In contrast, CNER offers a substantially richer label space (29 categories), with diverse supports and performance patterns, enabling a finer-grained and more interpretable analysis.
A full comparison of downstream effects, including the impact of splitting the \textsc{MISC} class into its CNER counterparts, is reported in Appendix~\ref{subsec:semantic-expressiveness}.

\paragraph{Manual validation}
The reliability of our evaluation framework critically depends on the accuracy of the semantic labels. 
We therefore conduct a quantitative analysis, manually annotating $30\%$ of the entire LitBank test set, and verifying the label assigned by CNER on gold clusters. This yields an overall $90\%$ precision, $87\%$ recall and $88\%$ $F_1$ score, confirming the high accuracy and reliability of our semantic label propagation observed in the qualitative evaluation presented in Appendix \ref{app:propagation_examples}.


\subsection{Interpretable Evaluation}
\label{sec:rq2}

We apply our semantic evaluation framework in order to identify semantic gaps in datasets that result in systematic model errors.

\paragraph{Category Distribution of Coreference Datasets.}
In Figure~\ref{fig:category-distributions}, we show the distribution of \textsc{CNER} semantic categories in OntoNotes, LitBank, and PreCo.
In LitBank, the category distribution produced by our labeling and propagation technique closely resembles the statistics reported by the dataset authors (cf. Section \ref{sec:exp-setup}), providing an external validation for the quality of our method.
At the same time, the distribution reveals the strong skew of LitBank annotations, a direct consequence of its restrictive annotation guidelines:
\textsc{PER} mentions are dominant, reflecting character-centered narratives, while categories such as \textsc{LOC} and \textsc{ORG} occur rarely.
Several semantic categories are entirely absent, including \textsc{MONEY}, \textsc{PLANT}, \textsc{ASSET}, \textsc{FOOD}, \textsc{DISCIPLINE}, \textsc{CELESTIAL}, \textsc{LANGUAGE}, \textsc{DISEASE}, \textsc{RELATION}, \textsc{LAW}, and \textsc{CULTURE}.
In contrast, OntoNotes and PreCo exhibit a more balanced distribution, with substantial coverage of \textsc{PER}, \textsc{ORG}, \textsc{LOC}, and \textsc{EVENT}, reflecting their multi-genre nature and less restrictive design choices.
These patterns illustrate a core challenge in out-of-domain evaluation that we will inspect in the following section: models trained on highly person-centric corpora such as LitBank may overfit to human-referential semantics, obtaining low performance compared to models trained more evenly across diverse entity types, as in OntoNotes and PreCo.

\paragraph{Model results.}

We evaluate the three variants of Maverick fine-tuned on OntoNotes, LitBank, and PreCo, and assess their performance both in-domain and out-of-domain, i.e., on the test set of the same dataset used for training, or the test set of a different one.
Table~\ref{tab:typed-macro}, left column, reports the macro-averaged Mention F$_1$ scores across all configurations, while we refer the reader to Appendix \ref{appendix:complete-evaluation} Table~\ref{tab:typed-micro} for a complete micro-averaged Mention F$_1$ score table.
We report that all models achieve strong in-domain performance.
However, when evaluated out-of-domain, Mention F$_1$ drops substantially, highlighting the difficulty of transferring mention extraction across datasets that differ in genre and annotation conventions.

A finer-grained view is reported in Figure~\ref{fig:per-class mentionF$_1$}, with per-class Mention F$_1$ scores for each model.
All models maintain strong in-domain performance across most semantic categories. In contrast out-of-domain results reveal pronounced discrepancies across both models and categories.
The LitBank-trained model, in particular, exhibits the lowest performance across multiple categories when evaluated on OntoNotes and PreCo.

To disentangle whether this result is due to the model predicting different span boundaries or wrongly predicting mentions of underrepresented semantic classes, we analyze Link F$_1$ scores using gold mentions as input (Table \ref{tab:typed-macro}, right column).
We report that the LitBank-trained model has lower performance in the out-of-domain setting, also when starting from gold mentions,  and refer the reader to Appendix~\ref{appendix:complete-evaluation} for a per-class analysis of Link F$_1$, where the performance drop on underrepresented classes is even more evident, confirming the concerns on person-centric bias of LitBank we raised in the previous section. In Appendix \ref{app:corss_domain_examples}, we also report a qualitative error analysis.
\begin{table}[t]
\centering
\small
\setlength{\tabcolsep}{6pt}
\renewcommand{\arraystretch}{1.05}

\begin{adjustbox}{max width=\columnwidth}
\begin{tabular}{
l r
@{\hspace{2.2em}}
c
@{\hspace{2.2em}}
l r
}
\toprule
\multicolumn{2}{c}{\textbf{Unrestricted}} &
&
\multicolumn{2}{c}{\textbf{Restricted}} \\
\cmidrule(lr){1-2}\cmidrule(lr){4-5}
Type & Count & & Type & Count \\
\midrule
\textsc{PER}        & 82 & & \textsc{PER}       & 69 \\
\textsc{EVENT}      & 65 & & \textsc{EVENT}     & 1  \\
\textsc{PSYCH}      & 54 & & \textsc{PSYCH}     & 1  \\
\textsc{PROPERTY}   & 52 & & \textsc{PROPERTY}  & 0  \\
\textsc{ARTIFACT}   & 47 & & \textsc{ARTIFACT}  & 9  \\
\textsc{MEDIA}      & 43 & & \textsc{MEDIA}     & 0  \\
\textsc{PLANT}      & 27 & & \textsc{PLANT}     & 0  \\
\textsc{MEASURE}    & 26 & & \textsc{MEASURE}   & 5  \\
\textsc{FEELING}    & 17 & & \textsc{FEELING}   & 0  \\
\textsc{LOC}        & 17 & & \textsc{LOC}       & 12 \\
\textsc{SUBSTANCE}  & 17 & & \textsc{SUBSTANCE} & 1  \\
\textsc{DISEASE}    & 16 & & \textsc{DISEASE}   & 0  \\
\textsc{GROUP}      & 16 & & \textsc{GROUP}     & 5  \\
\textsc{ASSET}      & 15 & & \textsc{ASSET}     & 0  \\
\textsc{ORG}        & 15 & & \textsc{ORG}       & 9  \\
\textsc{DATETIME}   & 14 & & \textsc{DATETIME}  & 0  \\
\textsc{LANGUAGE}   & 14 & & \textsc{LANGUAGE}  & 1  \\
\textsc{STRUCT}     & 13 & & \textsc{STRUCT}    & 8  \\
\textsc{PART}       & 12 & & \textsc{PART}      & 0  \\
\textsc{FOOD}       & 11 & & \textsc{FOOD}      & 0  \\
\textsc{DISCIPLINE} & 7  & & \textsc{DISCIPLINE}& 1  \\
\textsc{MONEY}      & 7  & & \textsc{MONEY}     & 0  \\
\textsc{CULTURE}    & 5  & & \textsc{CULTURE}   & 0  \\
\textsc{RELATION}   & 5  & & \textsc{RELATION}  & 0  \\
\textsc{CELESTIAL}  & 4  & & \textsc{CELESTIAL} & 3  \\
\textsc{LAW}        & 1  & & \textsc{LAW}       & 0  \\
\bottomrule
\end{tabular}
\end{adjustbox}

\caption{Comparison of mention counts between Un-\\restricted and Restricted settings across semantic types\\for our synthetic data.}
\label{tab:semantic-type-counts-aligned}
\end{table}

\begin{table*}[ht]
\centering
\small
\begin{tabular}{l ccc ccc ccc}
\toprule
 & \multicolumn{3}{c}{\textbf{CoNLL-F$_1$}} 
 & \multicolumn{3}{c}{\textbf{Link F$_1$ }} 
 & \multicolumn{3}{c}{\textbf{Mention F$_1$}} \\
\cmidrule(lr){2-4} \cmidrule(lr){5-7} \cmidrule(lr){8-10}
\textbf{Model}
 & \textbf{PreCo} & \textbf{Onto} & \textbf{Avg}
 & \textbf{PreCo} & \textbf{Onto} & \textbf{Avg}
 & \textbf{PreCo} & \textbf{Onto} & \textbf{Avg} \\
\midrule
maverick-mes-litbank
 & 45.5 & 51.7 & 48.6
 & 30.53 & 29.25 & 29.89
 & 34.40 & 26.75 & 30.58 \\

maverick-mes-litbank-augmented
 & 44.7 & 51.9 & 48.3
 & \textbf{34.96} & 26.37 & 30.67
 & 26.52 & \textbf{29.49} & 28.01 \\

maverick-mes-litbank-augmented-NR
 & \textbf{49.7} & \textbf{52.5} & \textbf{51.1}
 & \textbf{34.77} & \textbf{29.26} & \textbf{32.02}
 & \textbf{45.10} & \textbf{29.88} & \textbf{37.49} \\
\midrule
Diff. between augmented \& augmented-NR 
 &  &  & \textbf{+2.8}
 &  &  & \textbf{+1.35}
 &  &  & \textbf{+9.49} \\
\bottomrule
\end{tabular}
\caption{Comparative out-of-domain results of our three models used to test the benefit of targeted data augmentation, in terms of CoNLL-F1 and Macro Mention F$_1$ and Link F$_{1}$ scores on OntoNotes and PreCo.}
\label{tab:macro-results-compact}
\end{table*}

\subsection{Downstream Usage}
\label{sec:rq3}

\begin{figure}[t]
    \centering
    \includegraphics[width=\linewidth]{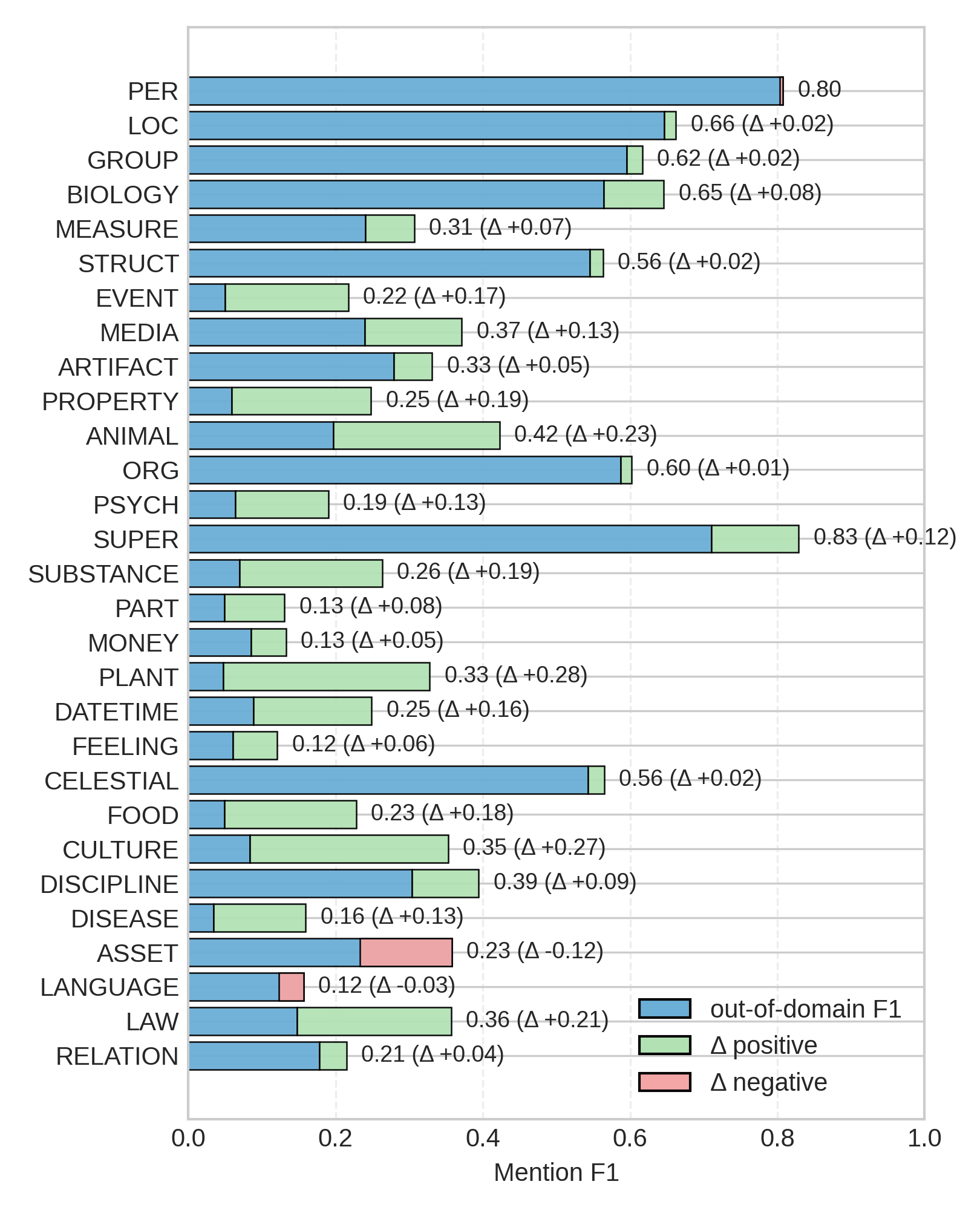}
    \caption{Performance difference between maverick-mes-litbank-NR and maverick-mes-litbank in terms of Mention F$_1$ in the out-of-domain setting. Positive values indicate improvements over the
    baseline.}
    \label{fig:litbank-mentionF1-unrestricted}
\end{figure}

We investigate whether our semantic diagnostics can be leveraged to improve out-of-domain performance through targeted, low-cost data augmentation for coreference resolution focused on underrepresented semantic categories.
Our goal is not to scale data augmentation, but to test whether semantically targeted diagnostics enable measurable improvements even with a very small amount of additional annotated data.

Specifically, as shown in the previous section, the LitBank-trained model exhibits a pronounced drop in out-of-domain coreference performance, particularly for semantic categories that were not annotated by design choice.
Motivated by this finding, we focus on the LitBank-trained model and evaluate whether adding a small number of documents in which those semantic categories appear can overcome these limitations.

We generate three synthetic narrative documents using GPT 5.1\footnote{\href{https://cdn.openai.com/gpt-5-system-card.pdf}{ChatGPT 5.1 System Card}}, each approximately 2{,}000 words long, written to match the LitBank narrative style while including mentions from semantic categories that are underrepresented or missing in the original training data (the prompt is provided in Appendix~\ref{appendix:Prompt}).
The documents are manually annotated under two alternative guidelines:
\begin{itemize}
    \item \textbf{Restricted annotation}, which restricts the annotation to entity types \textsc{PER}, \textsc{FAC}, \textsc{LOC}, \textsc{GPE}, \textsc{ORG}, and \textsc{VEH}, following the original LitBank guidelines and the directly associated classes that are present in the original CNER paper.
    \item \textbf{Unrestricted annotation}, which covers nominal and pronominal mentions to entities and concepts, without type-based restrictions.
\end{itemize}

In Table ~\ref{tab:semantic-type-counts-aligned}, we also report the distribution and support of semantic classes in our synthetic documents for both restricted and unrestricted annotation guidelines.

We train two augmented models on the LitBank training set by adding the three synthetic documents annotated under each guideline.
This yields two additional models:
(i) \texttt{maverick-mes-litbank-augmented}, trained with LitBank-like annotation; and
(ii) \texttt{maverick-mes-litbank-augmented-NR}, trained with unrestricted annotation (NR).\footnote{Training details are provided in Appendix~\ref{appendix:training_details}.}

In Table~\ref{tab:macro-results-compact} we show that the model trained with unrestricted annotation consistently outperforms both the LitBank baseline and the model augmented with LitBank-like annotations, improving +2.5 and +2.8  CoNLL-F$_1$ points, respectively.


These trends are reflected in the semantic evaluation metrics: Link F$_1$ shows a modest but consistent improvement, while Mention F$_1$ exhibits a substantially larger improvement of roughly 9.5 points.
This suggests that unrestricted annotation primarily benefits mention extraction, improving the model’s ability to detect and represent semantic categories that were previously underrepresented or absent.
A per-class analysis of out-of-domain Mention F$_1$ performance is presented in Figure ~\ref{fig:litbank-mentionF1-unrestricted}, and this further highlights per-class improvements.
Below, in Appendix~\ref{appendix:complete-evaluation}, we provide the Link F$_1$-based figures along with further analysis of their comparison.
We believe these results highlight the novel downstream capabilities of our semantic evaluation framework, which, unlike traditional aggregate statistical metrics, can be used directly to improve model performance. Not only that, we hope that future annotation campaigns will be able to leverage our evaluation framework to provide semantically diverse annotations, with the objective of training models that perform better out of domain.

\section{Conclusions}

In this work, we present a semantically-enhanced evaluation framework for coreference resolution that improves the interpretability of standard aggregate metrics. 
By overlaying Concept and Named Entity Recognition (CNER) onto coreference outputs, our two-step labeling and propagation technique densely annotates clusters with a fine-grained semantic categorization, enabling an analysis of mention extraction and coreference linking beyond what metrics such as CoNLL-F$_1$ can capture.
Using semantically typed Mention F$_1$ and Link F$_1$ scores, we exposed systematic domain-dependent weaknesses in current benchmarks and models, and showed how taking direct action to improve data quality leads to measurable gains in scores.

Overall, we demonstrate that semantic grounding of mentions enhances evaluation by addressing the interpretability gap and improving data quality, ultimately leading to improved model performance.

\section*{Limitations}

Despite the strengths of the proposed framework, several limitations should be acknowledged.
First, while we show that Concept and Named Entity Recognition (CNER) provides substantially higher coverage and finer-grained semantic categories than standard NER on coreference data, we do not perform a full intrinsic evaluation of the semantic propagation procedure used to assign cluster-level labels. In particular, due to the lack of manually annotated cluster-level semantic labels, we do not directly measure the accuracy of propagated semantic tags. As a result, the effect of semantic labeling errors on the proposed typed evaluation metrics is not explicitly quantified. We emphasize, however, that our goal is not to propose a state-of-the-art method for semantic annotation of coreference clusters, but rather to employ a simple and scalable baseline that enables semantic analysis of model behavior. We view this work as a step toward semantic coreference evaluation, and we hope it will motivate future efforts to construct high-quality, manually annotated benchmarks for this task.

Second, approximately 10\% of mentions remain unlabeled in each dataset. Qualitative inspection suggests that these predominantly correspond to clusters composed exclusively of pronouns, which cannot be directly tagged by CNER. While future work could potentially focus on handling these cases by training a lightweight classifier weakly supervised by the propagated labels, in this work, we restrict our analysis to the labeled subset of mentions.

Finally, the proposed technique relies on English-language CNER resources and has only been evaluated on English datasets. Extending the framework to other languages would require multilingual CNER models, which we leave for future work.

\section*{Acknowledgements}
\begin{center}
\noindent
    \begin{minipage}{0.1\linewidth}
        \begin{center}
            \includegraphics[scale=0.05]{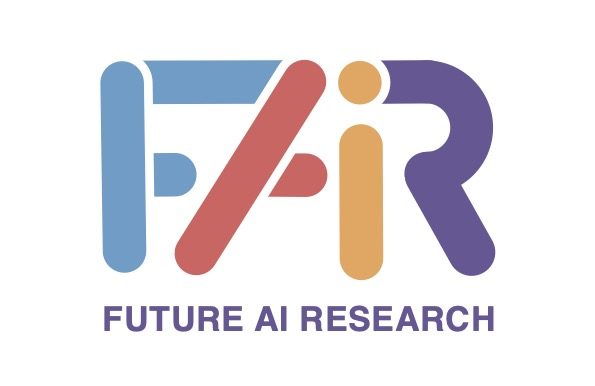}
        \end{center}
    \end{minipage}
    \hspace{0.01\linewidth}
    \begin{minipage}{0.70\linewidth}
         We gratefully acknowledge the support of the PNRR MUR project PE0000013-FAIR.
    \end{minipage}
    \hspace{0.01\linewidth}
    \begin{minipage}{0.1\linewidth}
        \begin{center}
            \includegraphics[scale=0.08]{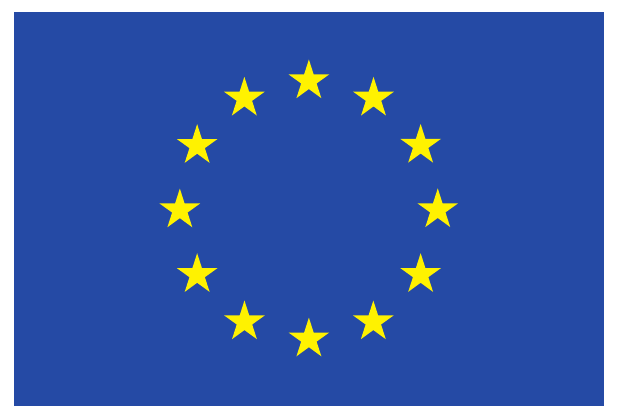}
        \end{center}
    \end{minipage}\\
\end{center}
\vspace{0.2cm}

We also gratefully acknowledge the support of the AI Factory IT4LIA project.

\bibliography{anthology,custom}

\appendix
\label{appendix}

\section{Additional Details on Effectiveness of our Labeling and Propagation Technique}
\label{app:propagation_heuristic}

In this Section we provide additional details on the process of evaluating our labeling and propagation technique.
We assess its effectiveness by i) measuring CNER mention coverage and ii) comparing CNER and standard NER as semantic annotation layers.

\begin{table*}[t]
\centering
\small
\renewcommand{\arraystretch}{1.15}
\setlength{\tabcolsep}{6pt}

\begin{tabular}{l l c c c c c c}
\hline
 &  & \multicolumn{3}{c}{\textbf{Overall Mentions}} &
      \multicolumn{3}{c}{\textbf{Pronoun Mentions}} \\
\cline{3-5} \cline{6-8}
\textbf{Dataset} & \textbf{System} &
\textbf{\%Dir} & \textbf{\%Prop} & \textbf{\%Any} &
\textbf{\%Dir} & \textbf{\%Prop} & \textbf{\%All} \\
\hline
OntoNotes & CNER & 53.27 & 35.80 & 89.07 & 0.00 & 83.47 & 83.47 \\
LitBank  & CNER & 37.50 & 50.53 & 88.03 & 0.00 & 84.09 & 84.09 \\
PreCo    & CNER & 71.36 & 15.57 & 86.94 & 0.00 & 62.46 & 62.46 \\
\hline
\end{tabular}

\caption{Mention coverage percentages for CNER, reported for overall mentions and pronominal
mentions. Direct coverage (\%Dir) corresponds to mentions labeled directly by the CNER
tagger, propagated coverage (\%Prop) via coreference links, and \%Any/\%All indicates total
labeled coverage.}
\label{tab:cner-coverage}
\end{table*}

\begin{table*}[t]
\centering
\small
\renewcommand{\arraystretch}{1.15}
\setlength{\tabcolsep}{6pt}

\begin{tabular}{l l c c c c c c}
\hline
 &  & \multicolumn{3}{c}{\textbf{Overall Mentions}} &
      \multicolumn{3}{c}{\textbf{Pronoun Mentions}} \\
\cline{3-5} \cline{6-8}
\textbf{Dataset} & \textbf{System} &
\textbf{\%Dir} & \textbf{\%Prop} & \textbf{\%Any} &
\textbf{\%Dir} & \textbf{\%Prop} & \textbf{\%All} \\
\hline
OntoNotes & WikiNEuRal & 21.65 & 31.13 & 52.77 & 0.00 & 47.71 & 47.71 \\
LitBank  & WikiNEuRal &  2.26 & 27.37 & 29.63 & 0.00 & 36.44 & 36.44 \\
PreCo    & WikiNEuRal & 12.86 &  9.98 & 22.84 & 0.00 & 28.90 & 28.90 \\
\hline
\end{tabular}

\caption{Mention coverage percentages for the NER-based baseline (WikiNEuRal), reported for
overall mentions and pronominal mentions. Direct coverage (\%Dir) corresponds to mentions
labeled via mention assignment, propagated coverage (\%Prop) via category propagation, and
\%Any/\%All indicates total labeled coverage.}
\label{tab:ner-coverage}
\end{table*}

\begin{table}[!ht]
\centering
\small
\setlength{\tabcolsep}{6pt}
\renewcommand{\arraystretch}{1.05}

\begin{adjustbox}{max width=\columnwidth}
\begin{tabular}{
l r r
@{\hspace{2.2em}}
c
@{\hspace{2.2em}}
l r r
}
\toprule
\multicolumn{3}{c}{\textbf{WikiNEuRal (NER)}} &
&
\multicolumn{3}{c}{\textbf{CNER}} \\
\cmidrule(lr){1-3}\cmidrule(lr){5-7}
Type & Link F$_1$ & Support & & Type & Link F$_1$ & Support \\
\midrule

\textsc{PER} & 0.896 & 46{,}147 & & \textsc{PER} & 0.890 & 58{,}153 \\
\textsc{LOC} & 0.849 & 12{,}988 & & \textsc{LOC} & 0.828 & 12{,}458 \\
\textsc{ORG} & 0.772 & 9{,}756  & & \textsc{ORG} & 0.761 & 8{,}645 \\
\textsc{MISC} & 0.823 & 7{,}096  & & \textsc{GROUP} & 0.854 & 5{,}222 \\
\addlinespace

& & & & \textsc{MEDIA} & 0.838 & 4{,}034 \\
& & & & \textsc{SUPER} & 0.888 & 3{,}782 \\
& & & & \textsc{ARTIFACT} & 0.723 & 2{,}291 \\
& & & & \textsc{EVENT} & 0.678 & 1{,}091 \\
& & & & \textsc{DATETIME} & 0.714 & 890 \\
& & & & \textsc{MEASURE} & 0.655 & 866 \\
& & & & \textsc{BIOLOGY} & 0.921 & 658 \\
& & & & \textsc{PROPERTY} & 0.894 & 839 \\
& & & & \textsc{PART} & 0.886 & 76 \\
& & & & \textsc{CELESTIAL} & 0.881 & 53 \\
& & & & \textsc{LAW} & 0.806 & 29 \\
& & & & \textsc{ASSET} & 0.542 & 29 \\
& & & & \textsc{DISCIPLINE} & 0.473 & 29 \\
& & & & \textsc{LANGUAGE} & 0.698 & 23 \\
& & & & \textsc{FOOD} & 0.722 & 18 \\
& & & & \textsc{CULTURE} & 0.800 & 15 \\
& & & & \textsc{FEELING} & 0.813 & 15 \\
& & & & \textsc{RELATION} & 0.621 & 15 \\
& & & & \textsc{DISEASE} & 0.588 & 10 \\
& & & & \textsc{PLANT} & 1.000 & 5 \\

\bottomrule
\end{tabular}
\end{adjustbox}

\caption{Distribution of semantic types with Link F$_1$ and support, comparing standard NER (WikiNEuRal) and fine-grained CNER. While NER aggregates diverse concepts under \textsc{MISC}, CNER redistributes them across semantically specific categories, providing both broader coverage and improved interpretability.}
\label{tab:ner-vs-cner-full}
\end{table}

\subsection{Mention Coverage Additional Details}
In Table~\ref{tab:cner-coverage} we see the details of CNER mention coverage, with an additional focus on pronominal mentions.
As expected the percentage is lower than labeled mentions in general, which hints that, as we hypothesized, the gap of roughly 10\% of unlabeled mentions is mostly composed of pronominal clusters.

\subsection{NER Mention Coverage and Pronoun Propagation}

Table~\ref{tab:ner-coverage} reports detailed statistics on mention labeling when
using a NER-based baseline (WikiNEuRal). The table distinguishes between labels assigned
directly by the tagger and those obtained via coreference-based propagation, and reports
results separately for all mentions and for pronominal mentions.

From Table~\ref{tab:ner-coverage}, we observe the following trends. First, the
proportion of unlabeled pronominal mentions is substantially higher than that of
non-pronominal mentions, mirroring the overall limited coverage of the NER-based baseline.
Second, while label propagation partially mitigates this issue, the percentage of
pronominal mentions receiving a semantic label remain relatively low across all datasets.



\subsection{Semantic Granularity of NER vs.\ CNER}
\label{subsec:semantic-expressiveness}

Beyond raw mention coverage, we also analyze the semantic granularity of NER when used as a semantic layer for evaluation. In this subsection, we report a complete comparison between semantic evaluation conducted with a standard NER-based layer and with CNER as the semantic layer.

Table \ref{tab:ner-vs-cner-full} shows that using CNER enables a
substantially richer semantic label space, comprising 29 entity classes, compared to the more limited tagset typically provided by NER, also showing Link F$_1$ as a performance measure. We show that more diverse class supports reveal finer-grained performance patterns that are not observable while using NER, which brings scores to be aggregated under the MISC class.

\begin{figure*}[t]
    \centering
    \includegraphics[width=0.6\linewidth]{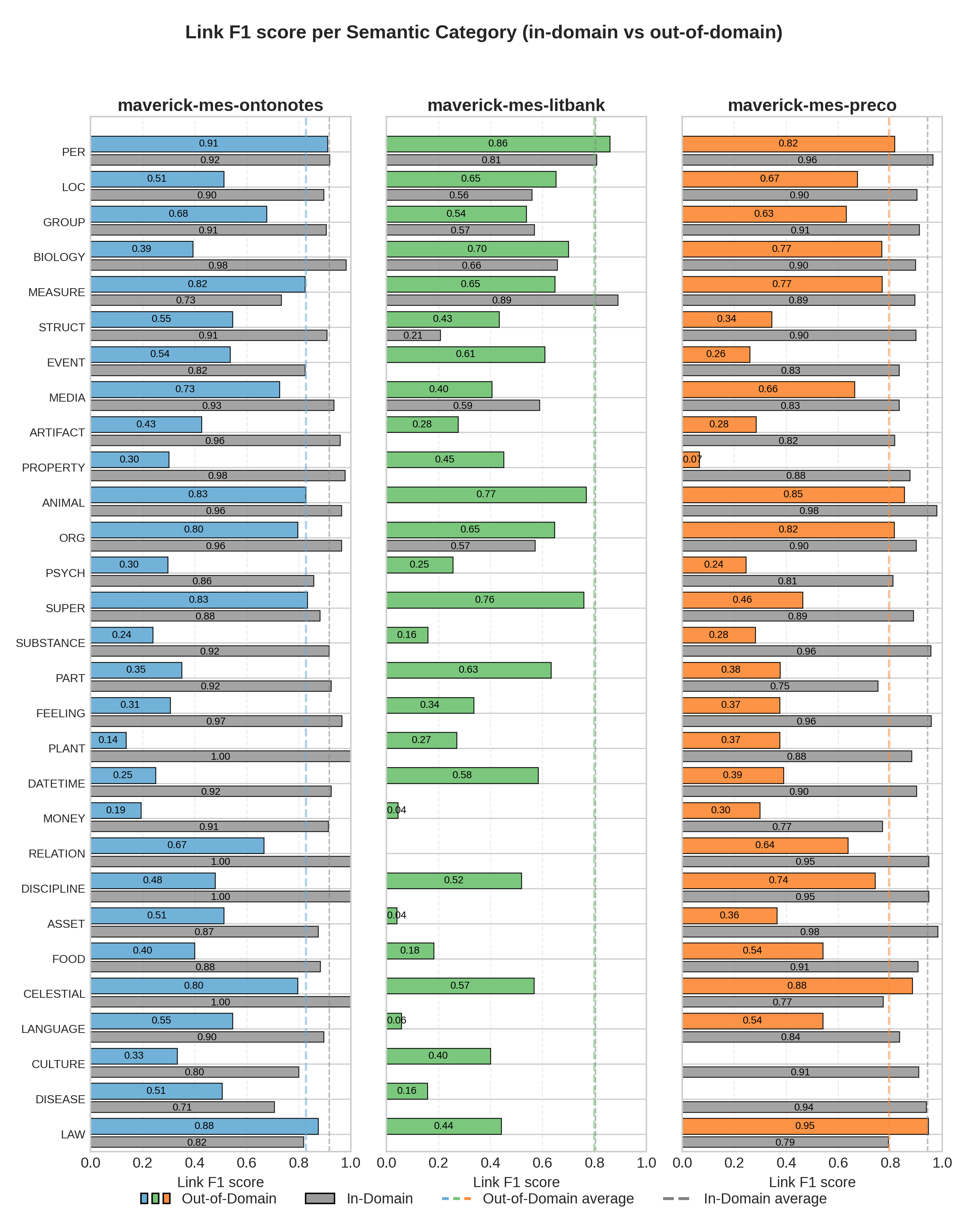}
    \caption{Per-class Link F$_1$ scores for maverick-mes-ontonotes,
    maverick-mes-litbank, and maverick-mes-preco. Grey bars indicate
    in-domain performance, while colored bars indicate out-of-domain performance. Classes
    are ordered by decreasing support in the LitBank dataset.}
    \label{fig:linkF$_1$-by-model}
\end{figure*}

\section{Complete Evaluation Results}
\label{appendix:complete-evaluation}

In this Section, we report additional evaluation results, omitted from the main paper for clarity and space constraints.

We first present the Micro-averaged results of our evaluation on maverick-mes-ontonotes, maverick-mes-litbank and maverick-mes-preco, then complement the per-class Mention F$_1$ evaluation presented in Section \ref{sec:rq2} with Link F$_1$ scores.


\subsection{Micro-averaged Results}
Table \ref{tab:typed-micro} reports the Micro F$_1$ scores for the analyzed models.
We observe a decline in out-of-domain performance on Mention F$_1$ scores for all models, with a steep decrease in maverick-mes-litbank as it reaches the lowest score on Ontonotes and ties with maverick-mes-ontonotes on the PreCo dataset.
In Link-F$_1$, we observe maverick-mes-ontonotes outperforming maverick-mes-litbank on its own domain.

\begin{table*}[ht]
\centering
\small
\begin{tabular}{l ccc ccc}
\toprule
 & \multicolumn{3}{c}{\textbf{Mention F$_{1}$}} 
 & \multicolumn{3}{c}{\textbf{Link F$_{1}$}} \\
\cmidrule(lr){2-4} \cmidrule(lr){5-7}
\textbf{Model} 
 & \textbf{OntoNotes} & \textbf{LitBank} & \textbf{PreCo}
 & \textbf{OntoNotes} & \textbf{LitBank} & \textbf{PreCo} \\
\midrule
maverick-mes-ontonotes & \textbf{0.90} & 0.74 & 0.49 & \textbf{0.92} & \textbf{0.89} & 0.77 \\
maverick-mes-litbank   & 0.67 & \textbf{0.92} & 0.50 & 0.80 & 0.80 & 0.82 \\
maverick-mes-preco     & 0.70 & 0.64 & \textbf{0.93} & 0.82 & 0.77 & \textbf{0.94} \\
\bottomrule
\end{tabular}
\caption{Micro Mention F1 and Link F1 results for each of the comparison systems on all our datasets. }
\label{tab:typed-micro}
\end{table*}

\subsection{Link F$_1$}
Figure~\ref{fig:linkF$_1$-by-model} reports per-class Link F$_1$ scores for our comparison systems. For each model we show in-domain performance in grey and out-of-domain performance in color. Semantic classes are ordered by decreasing support in the LitBank dataset.

The results highlight several consistent patterns. In line with the quantitative results reported in the main corpus, the LitBank-trained model exhibits poor out-of-domain performance, particularly on semantic categories with low or zero support in the LitBank training set due to its restrictive annotation guidelines. Notably, several classes are entirely absent from LitBank, leading to near-zero performance when evaluated out of domain.
Moreover, even in-domain, maverick-mes-litbank underperforms compared to maverick-mes-ontonotes. As reported in Table \ref{tab:typed-micro}, the LitBank model reaches an in-domain Link F$_1$ of 80, which is lower than the corresponding score achieved by the OntoNotes-trained model. This further suggests that limited semantic coverage negatively affects both generalization and in-domain linking quality.
A clear performance decrease is also observable on the OntoNotes and PreCo trained models from in-domain to out-of-domain performance, especially in classes such as \textsc{SUBSTANCE}, \textsc{PART}, \textsc{FEELING}, \textsc{PLANT}, \textsc{DATETIME} and \textsc{MONEY}, where we have a decrease that is often more than 60 Link F$_1$ points.

\section{Qualitative Error Analysis}
\label{semantic-result-qualitative}

This section provides qualitative examples illustrating the behavior of the proposed semantic propagation framework and the types of semantic and linking coreference errors revealed by our evaluation. The examples complement the quantitative results by offering concrete illustrations of domain-specific patterns discussed in the main corpus.

\subsection{CNER labeling and propagation examples.}
\label{app:propagation_examples}
In Table \ref{table:propagation}, we show two qualitative examples of CNER labeling and propagation onto coreference clusters.

\begin{table*}[ht]
\centering
\small
\resizebox{\linewidth}{!}{%

\begin{tabular}{p{0.17\textwidth}|p{0.83\textwidth}}
\hline\hline
\textbf{Stage} & \textbf{Annotation} \\
\hline
\hline
Example 1 
\\
\hline
\hline
CNER & While all this is going on, Mr. \corefBlack{Clinton}{PER}  is overseas. \corefBlack{President Clinton}{PER} was in \corefBlack{Northern Ireland}{LOC} when he heard the \corefBlack{Supreme Court}{ORG} \corefBlack{decision}{EVENT} . He talked to \corefBlack{Al Gore}{PER} on the \corefBlack{phone}{ARTIFACT} from \corefBlack{Belfast}{LOC}. This is Mr. \corefBlack{Clinton}{PER} 's  \corefBlack{third}{MEASURE} \corefBlack{visit}{EVENT} to \corefBlack{Northern Ireland}{LOC}"
\\
\hline
Coreference & "While all this is going on, \corefBlack{Mr. Clinton}{1} is overseas. \corefBlack{President Clinton}{1} was in \corefBlack{Northern Ireland}{2} when \corefBlack{he}{1} heard the Supreme Court decision.  \corefBlack{He}{1} talked to Al Gore on the phone from Belfast. This is \corefBlack{Mr. Clinton}{1} 's third visit to \corefBlack{Northern Ireland}{2}"
\\
\hline
(1) Mention Assignment & "While all this is going on, \corefBlue{Mr. Clinton}{1-PER} is overseas.  was in \corefRed{Northern Ireland}{2-LOC} when \corefBlack{he}{1-Unassigned} heard the Supreme Court decision .  \corefBlack{He}{1-Unassigned} talked to Al Gore on the phone from Belfast. This is \corefBlue{Mr. Clinton}{1-PER} 's third visit to \corefRed{Northern Ireland}{2-LOC}"
\\
\hline
(2) Cluster Propagation & " While all this is going on, \corefBlue{Mr. Clinton}{1-PER}  is overseas. \corefBlue{President Clinton}{1-PER} was in \corefRed{Northern Ireland}{2-LOC} when \corefBlue{he}{1-PER} heard the Supreme Court decision .  \corefBlue{He}{1-PER} talked to Al Gore on the phone from Belfast. This is \corefBlue{Mr. Clinton}{1-PER} 's third visit to \corefRed{Northern Ireland}{2-LOC}"
\\
\hline
\hline
Example 2
\\
\hline
\hline
CNER &  "\corefBlack{Tomorrow}{DATETIME}'s  \corefBlack{summit}{EVENT}  \corefBlack{meeting}{EVENT} will bring  \corefBlack{Ehud Barak}{PER} and  \corefBlack{Yasser Arafat}{PER} to the \corefBlack{resort city}{STRUCT} of \corefBlack{Sharm El - Sheikh}{LOC}. Getting both to attend was not an easy task."
\\
\hline
Coreference &  "Tomorrow's summit meeting will bring \corefBlack{\corefBlack{Ehud Barak}{1} and  \corefBlack{Yasser Arafat}{2}}{3} to the resort city of Sharm El-Sheikh. Getting \corefBlack{both}{3} to attend was not an easy task."
\\
\hline
(1) Mention Assignment &  "Tomorrow's summit meeting will bring \corefOlive{\corefRed{Ehud Barak}{1-PER} and \corefBlue{Yasser Arafat}{2-PER}}{3-PER} to the resort city of Sharm El-Sheikh. Getting \corefBlack{both}{3-Unassigned} to attend was not an easy task."
\\
\hline
(2) Cluster Propagation &  "Tomorrow's summit meeting will bring \corefOlive{\corefRed{Ehud Barak}{1-PER} and  \corefBlue{Yasser Arafat}{2-PER}}{3-PER} to the resort city of Sharm El-Sheikh. Getting \corefOlive{both}{3-PER} to attend was not an easy task."
\\
\hline

\hline

\end{tabular}
}
\caption{Propagation examples annotation.}
\label{table:propagation}
\end{table*}

In the first example, from a news-domain text in OntoNotes, CNER labeling and propagation ensure type consistency across all mentions within the same entity cluster.  
In the first row, we show the high coverage of named entities and concepts as CNER annotated spans, while in the second row, we report the different span boundaries found in the gold annotated coreferences of OntoNotes.
During the Mention Assignment stage, we can assign to every non-pronoun mention a CNER label, and with cluster propagation, we can obtain complete coverage of coreference mentions.

The same happens in the second example, in which we can propagate our CNER labels for plural references.
In fact, CNER annotation assigns PER to the named individuals Ehud Barak and Yasser Arafat, and the Mention Assignment step tags them correctly as PER, leaving out plural references, which are inherited in the following Cluster Propagation step, demonstrating the technique’s ability to extend semantic categories across implicitly linked mentions.
\subsection{Interpretable Error Examples.}
\label{app:corss_domain_examples}
Table~\ref{table:models-qualitative} presents qualitative examples illustrating cross-domain discrepancies among our Maverick-mes models trained on LitBank, OntoNotes, and PreCo. These examples highlight systematic differences in mention boundary detection, semantic category coverage, and cross-type linking behavior.
\begin{table*}[!ht]
\centering
\small
\resizebox{\linewidth}{!}{%

\begin{tabular}{p{0.12\textwidth}|p{0.88\textwidth}}
\hline\hline
\textbf{Model} & \textbf{Output} \\
\hline
\hline
Ex. 1 & \textit{maverick-mes-litbank does not recognize CULTURE mentions.}
\\
\hline
\hline
maverick-litbank & "... And \corefPurple{people}{singleton-GROUP} have different opinions about the movement. \corefOlive{Some}{singleton-GROUP} think street art is a crime and destroys property. But \corefBlack{others}{singleton-GROUP} see this art as a rich form of non-traditional cultural expression. \corefBlue{Many experts}{singleton-GROUP} say the movement began in \corefGreen{New York City}{1-LOC} in the nineteen sixties..."
\\
\hline
maverick-ontonotes &... And people have different opinions about \corefPurple{the movement}{2-CULTURE}. Some think \corefRed{street art}{3-MEDIA} is a crime and destroys property. But others see \corefRed{this art}{3-MEDIA} as a rich form of non-traditional cultural expression. Many experts say \corefPurple{the movement}{2-CULTURE} began in \corefGreen{New York City}{4-LOC} in the nineteen sixties..."
\\
\hline
\hline
Ex. 2 & \textit{maverick-mes-litbank does not recognize \textsc{DISEASE} mentions}.
\\
\hline
\hline
maverick-litbank & "... The insects carry serious diseases like malaria. It is estimated that almost 630,000 people died from malaria and malaria-related causes in 2012, and most of these cases were in African countries..."
\\
\hline
maverick-ontonotes & "... \corefPurple{The insects}{0-BIOLOGY} carry serious diseases like \corefRed{malaria}{1-DISEASE}. It is estimated that almost 630,000 people died from \corefRed{malaria}{1-DISEASE} and \corefRed{malariarelated}{1-DISEASE} causes in 2012, and most of these cases were in African countries..."
\\
\hline
\hline
Ex. 3 & \textit{Using gold mentions, maverick-mes-litbank is unable to cluster \textsc{MONEY}.}
\\
\hline
\hline
maverick-litbank & "... \corefPurple{With the previous 800 yuan income tax threshold}{singleton-MONEY}, if \corefBlue{it}{singleton-Unassigned} were strictly enforced, 80\% of beggars would have to pay..."
\\
\hline
maverick-preco & " ... With \corefPurple{the previous 800 yuan income tax threshold}{6-MONEY}, if \corefPurple{it}{6-MONEY} were strictly enforced, 80\% of beggars would have to pay..."
\\
\hline
\hline
Ex. 4 & \textit{Using gold mentions, maverick-mes-litbank is unable to cluster \textsc{MONEY}.}
\\
\hline
\hline
maverick-litbank & "...When \corefPurple{he}{7-PER} came home, \corefPurple{he}{7-PER} said, ‘Call \corefRed{those servants who have \corefPurple{my}{7-PER} money}{10-PER}...’"
\\
\hline
maverick-preco & "...When \corefPurple{he}{8-PER} came home, \corefPurple{he}{8-PER} said, ‘Call \corefRed{those servants who have \corefGreen{\corefPurple{my}{8-PER} money}{15-MONEY}}{10-PER}...’"
\\
\hline
\hline
Ex. 5 & \textit{Using gold mentions, maverick-mes-litbank is unable to cluster \textsc{DISEASE}.}
\\
\hline
\hline
maverick-litbank & "...\corefPurple{Sharon Osbourne, \corefGreen{Ozzy’s}{0-PER} long-time manager, wife and best friend}{1-PER}, announced to the world that \corefPurple{she}{1-PER} ’d been diagnosed with colon cancer. ... \corefPurple{Sharon}{1-PER} announced in April that the cancer was in remission, but just weeks after that announcement, \corefRed{son Jack}{5-PER} entered drug and alcohol rehabilitation in California..."
\\
\hline
maverick-preco & "...\corefPurple{Sharon Osbourne, \corefGreen{Ozzy’s}{0-PER} long-time manager, wife and best friend}{1-PER}, announced to the world that \corefPurple{she}{1-PER} ’d been diagnosed with \corefRed{colon cancer}{3-DISEASE}. ... \corefPurple{Sharon}{1-PER} announced in April that \corefRed{the cancer}{3-DISEASE} was in remission, but just weeks after that announcement, \corefBlue{son Jack}{4-PER} entered drug and alcohol rehabilitation in California..."
\\
\hline
\end{tabular}
}
\caption{Representative cross-domain annotation errors. Each example contrasts the LitBank-trained model with OntoNotes or PreCo models, showing domain-specific gaps in semantic coverage and mention linking.}
\label{table:models-qualitative}
\end{table*}
In the first two examples, we observe mention extraction failures in the LitBank-trained model, particularly for underrepresented categories such as \textsc{CULTURE} and \textsc{DISEASE}.
In Example~1, the LitBank model identifies only generic \textit{Group} mentions (e.g., “people,” “some,” “others”), whereas conceptual entities like \textit{the movement} remain unlabeled. In contrast, the OntoNotes model correctly extracts these and assigns \textsc{Culture} and \textsc{Media} categories, reflecting broader semantic coverage.

In Example~2, the LitBank model omits all disease-related mentions, while OntoNotes successfully labels \textit{malaria} and related expressions, as well as biological entities such as \textit{the insects}. This underscores OntoNotes’ stronger lexical diversity and its inclusion of scientific and factual content, compared to LitBank’s narrative focus.

Examples~3 and~4 highlight linking errors involving financial entities. Even with gold mentions, the LitBank model fails to link money-related expressions, treating them as singletons, whereas the PreCo model clusters them accurately and correctly handles possessive references (e.g., \textit{my money}).

Finally, Example~5 demonstrates that the LitBank-trained model fails to link disease mentions (\textit{colon cancer}, \textit{the cancer}), due to the scarcity of such categories in LitBank. The PreCo model, by contrast, clusters them consistently.

Overall, these qualitative examples reveal that LitBank-trained models struggle with non-human and abstract referents, such as cultural, financial, and biomedical entities, due to LitBank’s narrative bias and limited exposure to factual or technical domains. This domain imbalance results in reduced mention recall and weaker cross-type linking generalization.

\section{Data Augmentation Experiment Details}
\label{appendix:Prompt}

In this Section, we provide a more detailed analysis of the data augmentation experiments. 
Specifically, we report: i) the training setup and hyperparameters of the models, and 
ii) details about the newly generated training data used in our experiments.

\subsection{Training Details}
\label{appendix:training_details}

All models were trained using the official Maverick codebase%
\footnote{\url{https://github.com/SapienzaNLP/maverick-coref}}.
Training was performed on a single NVIDIA RTX 4090 GPU with 24GB of VRAM, using CUDA~12.4.

We focus on the maverick-mes architecture, employing a \textsc{DeBERTa-v3-large}
encoder as the underlying language model. The encoder was fine-tuned during training (i.e.,
not frozen), and span representations were constructed by concatenating the start and end
token embeddings.

The most relevant hyperparameters are reported
below:

\begin{itemize}
    \item Optimizer: Adafactor
    \item Learning rate: $3 \times 10^{-5}$
    \item Weight decay: $0.01$
    \item Warm-up steps: $6{,}000$
    \item Total training steps: $8{,}000$
    \item Gradient accumulation: $4$ steps
    \item Gradient clipping: $1.0$
    \item Precision: FP16
\end{itemize}

Training was conducted using PyTorch Lightning with a deterministic setup (random seed
set to~30). Validation was performed twice per epoch. Early stopping was enabled based on
the CoNLL-2012 F$_1$ validation score, with a patience of 120 validation checks. Model
checkpointing was performed using the same metric.

\subsection{Training Data Details}

\begin{table*}[!ht]
\centering

\begin{tcolorbox}[
  colback=gray!6,
  colframe=gray!6,
  boxrule=0pt,
  arc=2pt,
  left=6pt,right=6pt,top=6pt,bottom=6pt
]
\textbf{Synthetic Document Generation Prompt.}

We aim to generate from scratch a long-form document written in the style of texts found in the LitBank coreference benchmark, a well-known dataset for long-document coreference resolution.

\textbf{Document requirements:}
\begin{itemize}[leftmargin=*, itemsep=1pt, topsep=2pt, parsep=0pt]
  \item \textbf{Length:} approximately 2{,}000 words.
  \item \textbf{Style:} narrative, cohesive, and information-dense, resembling LitBank benchmark documents.
  \item \textbf{Structure:} suitable for coreference resolution, with recurring entities, implicit references, and long-range dependencies.
\end{itemize}

The text must naturally and meaningfully include all of the following CNER semantic classes, with multiple occurrences distributed across the document.

\textbf{Required semantic classes:}

$$[...]$$

All semantic classes must be embedded organically within the narrative rather than enumerated explicitly in the generated text, and the overall tone may be historical, academic, or narrative, provided it remains consistent with the LitBank document style and supports complex, long-distance coreference phenomena.
\end{tcolorbox}
\caption{Prompt used for synthetic document generation; the list of required semantic classes includes class name, description, and examples.}
\label{tab:synthetic-prompt}
\end{table*}

In this subsection, we describe the prompt used to generate the additional training data
employed in the data augmentation experiments.

The data was generated using GPT 5.1\footnote{\url{https://cdn.openai.com/gpt-5-system-card.pdf}}. The model was prompted to produce synthetic training
examples consistent with the target task formulation. The exact prompt used for data
generation is reported in Table \ref{tab:synthetic-prompt}.

\paragraph{Annotation Details}

The generated texts were annotated according to two different annotation schemes:
i) \emph{unrestricted annotation}, and
ii) \emph{restricted annotation}, inspired by the LitBank annotation guidelines.

In the restricted setting, annotations were designed to mirror the LitBank scheme, which
does not annotate all semantic classes but is limited to the following entity types:
\textsc{PERSON} (PER), \textsc{FACILITY} (FAC), \textsc{LOCATION} (LOC),
\textsc{GEO-POLITICAL ENTITY} (GPE), \textsc{ORGANIZATION} (ORG), and
\textsc{VEHICLE} (VEH).
These entity types have a direct map with CNER classes, as described in the original Concept and Named Entity Recognition paper \citep{martinelli-etal-2024-cner}.

\begin{figure*}[t]
    \centering
    \begin{minipage}[t]{0.48\linewidth}
        \centering
        \includegraphics[width=\linewidth]{MentionsF1_litbank_litbankNR_new.png}
        \caption{Delta of out-of-domain Mention F$_1$ for maverick-mes-litbank-NR,
        compared to maverick-mes-litbank. Positive values indicate improvements over the
        baseline.}
        \label{fig:litbank-mentionF1-unrestricted}
    \end{minipage}\hfill
    \begin{minipage}[t]{0.48\linewidth}
        \centering
        \includegraphics[width=\linewidth]{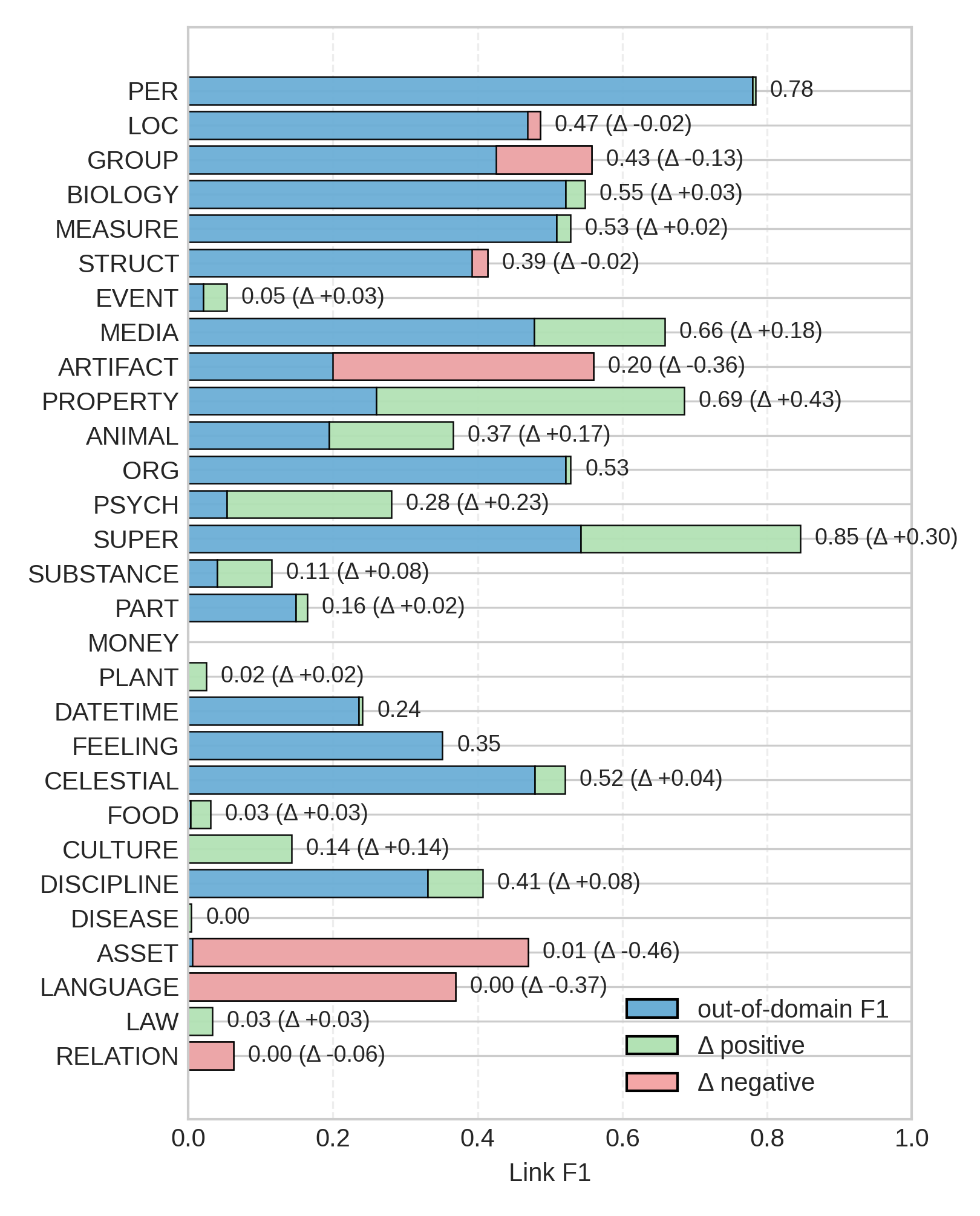}
        \caption{Delta of out-of-domain Link F$_1$ for maverick-mes-litbank-NR,
        compared to maverick-mes-litbank. Positive values indicate improvements over the
        baseline.}
        \label{fig:litbank-linkF1-unrestricted}
    \end{minipage}
\end{figure*}

\subsection{Evaluation Results on LitBank Models}
\label{subsec:litbank-results}

In this subsection, we report the per-class results of our evaluation framework applied to
Maverick-mes models trained on data augmented with generated texts.
We focus on out-of-domain performance, and we analyze the
impact of unrestricted versus restricted annotation schemes, to complement the results obtained with a fine grained per-class evaluation.

Figure~\ref{fig:litbank-mentionF1-unrestricted} reports the delta in Mention F$_1$ scores, between a maverick-mes-litbank-augmented-NR and maverick-mes-litbank. We observe a clear improvement
in Mention F$_1$ for nearly all entity classes, with particularly strong gains for classes that are underrepresented or entirely absent in the original LitBank training data. 

Figure~\ref{fig:litbank-linkF1-unrestricted} reports the corresponding results for
out-of-domain Link F$_1$, comparing the same augmented model against the baseline.

In contrast to the mention-level results, improvements in Link F$_1$ are more limited.
This mirrors the behavior already observed in the main paper. A possible explanation is a mismatch in the distribution of coreference links between the generated data and the test sets, for instance, due to an over-representation of singleton mentions in the augmented
data. A more detailed analysis of this phenomenon is left for future work.

\section{CNER Categories}
\label{app:cner}

In this Section, we provide a detailed description of the CNER categories used in our
experiments. We include this information for completeness, since CNER is employed as the
semantic layer of our model. A clear understanding of the entity and concept types covered by each category is essential both for interpreting our results and for enabling potential users of our evaluation framework to conduct more informed analyses.

Table~\ref{tab:cner-categories} reports the full list of CNER categories, together with their
definitions and representative examples.

\begin{table*}[t]
\centering
\footnotesize
\renewcommand{\arraystretch}{1.05}
\begin{tabularx}{\textwidth}{@{} l X X @{}}
\toprule
\textbf{Category} & \textbf{Description} & \textbf{Examples} \\
\lightrule

ANIMAL & Living beings (excluding humans) with the ability to move and perceive their surroundings. &
dog, cat, mammal, carnivore, brown bear, African Wild Dog, Great White Shark \\

\lightrule
ARTIFACT & All the objects, artifacts, tools, products and items &
vehicle, software, mouse, data stream, Windows XP, Fiat Panda \\

\lightrule
ASSET & Assets, resources, or possessions with economic or intrinsic value. &
capital, stock, wealth, resource, phone bill, Federal Perkins Loan, Investment in Russia \\

\lightrule
BIOLOGY & Biological entities, including living organisms, cells, or biological components &
protein, cell, living organism, lipid, Herpes Simplex Virus, Escherichia Coli \\

\lightrule
CELESTIAL & celestial bodies as Planets, stars, asteroids, galaxies and other astronomical objects. &
comet, nebulae, Sun, Neptune, Asteroid 187 Lamberta, Proxima Centauri \\

\lightrule
CULTURE & Cultural aspects, traditions, customs, and practices associated with specific groups or societies. &
religion, feminism, socialism, capitalism, anarchism, doctrine, cult, Islam, Buddhism \\

\lightrule
DATETIME & Dates and times &
18 March, Saturday, 1979, the evening of 19 November, 15:30 am \\

\lightrule
DISEASE & medical conditions, illnesses, disorders, and health-related issues affecting living organisms. &
infection, allergy, metastasis, complication, acne, Alzheimer’s Disease, Cystic Fibrosis \\

\lightrule
DISCIPLINE & specific fields of study, knowledge, or expertise. It includes academic disciplines, areas of research, and professional domains. &
discipline, sport, football, computer science, anatomy, long jump. \\

\lightrule
EVENT & Events, phenomenon or activities that occur at specific times or places. It includes both significant and everyday occurrences &
crime, professorship, temperature change, 2003 Wimbledon Championships, Cannes Film Festival. \\

\lightrule
FEELING & Emotions, sensations, and subjective experiences related to human or animal consciousness. &
affection, attachment, agitation, craving, urge, temptation. \\

\lightrule
FOOD & edible items, dishes, beverages, and culinary products that are consumed for nourishment or enjoyment &
beverage, dish, pork, lasagna, Carbonara, Sangiovese, Cheddar Beer Fondue, Pizza Margherita. \\

\lightrule
GROUP & group of people or animals &
staff, social group, panel, militia, community, trio, duo, family, genealogy, alliance, nationality, peoples \\

\lightrule
LANGUAGE & individual language-related items, such as words, phrases, or idiomatic expressions &
discourse, context, lexeme, morpheme, appellation, eponym, nickname, vowel, syllable, headword \\

\lightrule
LAW & legal principles, regulations, and rules governing society and various aspects of life &
law, civil law, administrative law, martial law, shariah, ordinance, civil right, Magna Carta, Islamic Law \\

\lightrule
LOC & geographical locations, such as villages, towns, cities, regions, countries, continents, landmarks, or natural features &
space, surface, street, road, town, Rome, Lake Paiku, Mississippi River. \\

\lightrule
MEASURE & units of measurement and quantification used to determine the size, quantity, or quality of various objects or phenomena. &
day, microsecond, millisecond, two, 35, 45\%, first, temperature, length \\

\lightrule
MEDIA & various forms of communication and entertainment media, such as newspapers, television shows, movies, social media or digital content. &
soundtrack, report, publication, language, English, Forbes, American Psycho \\

\lightrule
MONEY & monetary units, currencies, and financial values used in different contexts &
monetary unit, dollar, 15 euros, 1116 CHF \\

\lightrule
ORG & organizations, institutions, and companies involved in diverse sectors or activities &
Industry, commercial enterprise, San Francisco Giants, Google, Democratic Party. \\

\lightrule
PART & individual components or sections of larger entities or objects &
finger, chin, head, tail, femur, airplane wing, airplane’s wings, flower’s stem \\

\lightrule
PER & individuals or persons, including real people and historical figures &
doctor, historian, professor, musician, Ray Charles, Jessica Alba \\

\lightrule
PLANT & Types of trees, flowers, and other plants, including their scientific names. &
grass, peach tree, Forsythia, Artemisia Maritima. \\

\lightrule
PROPERTY & properties or attributes of objects, entities, or concepts &
thickness, height, dimension, shape, age \\

\lightrule
PSYCH & psychological concepts, mental states, and phenomena related to the human mind and behavior &
psychological feature, cognition, attention, necessity \\

\lightrule
RELATION & relationships, connections, and associations between entities or concepts &
apport, competition, comparison, bridge, relatedness, parentage, function, parity, transitivity \\

\lightrule
STRUCT & physical structures, including buildings, architectural designs, and engineered constructions made by humankind &
shelter, gravestone, refuge, tent, loft, San Peter’s Church, Golden Bridge \\

\lightrule
SUBSTANCE & chemical substances &
acid, bactericide, carbonyl, explosive, fertilizer, Zyclon B \\

\lightrule
SUPER & Mythological and religious entities. &
Apollo, Persephone, Aphrodite, Saint Peter, Pope Gregory I, Hercules. \\

\bottomrule
\end{tabularx}
\caption{Label, description, and instance examples of each of our CNER categories.}
\label{tab:cner-categories}
\end{table*}
\end{document}